\documentclass[11pt]{article}
\usepackage{multirow}
\usepackage[final]{acl}
\usepackage{colortbl}
\usepackage{times}
\usepackage{latexsym}
\usepackage{booktabs}
\usepackage[T1]{fontenc}
\usepackage{float}
\usepackage[utf8]{inputenc}
\usepackage{bm}
 \usepackage{amsmath} 
\usepackage{tikz} 
\usepackage{colortbl}
\usepackage[table]{xcolor}
\usepackage{pgf} 
\usepackage{microtype}
\usepackage[table]{xcolor}
\usepackage{booktabs}
\usepackage{tabularx}
\usepackage{multirow}
\definecolor{AcadBlue}{RGB}{220,230,242}
\usepackage[most]{tcolorbox} 
\usepackage{enumitem}        

\newtcolorbox{rubricitem}[1]{
    colback=AcadBlue!15,      
    colframe=AcadBlue!100,    
    coltitle=black,           
    title={#1},               
    fonttitle=\bfseries,      
    boxrule=0.8pt,            
    arc=3pt,                  
    left=4pt, right=4pt,      
    top=4pt, bottom=4pt,
    parbox=false,             
    breakable                 
}

\definecolor{ProprietaryBlue}{RGB}{220, 230, 242} 
\definecolor{OpenSourceGreen}{RGB}{220, 242, 230} 

\newcommand{\propcell}[1]{%
    \cellcolor{ProprietaryBlue!#1}{#1}
}
\usepackage{xcolor}
\usepackage{ragged2e}

\newcolumntype{L}{>{\raggedright\arraybackslash}p{0.15\textwidth}}  
\newcolumntype{V}{>{\raggedright\arraybackslash}p{0.08\textwidth}}  
\newcolumntype{C}{>{\raggedright\arraybackslash}X}                 
\newcolumntype{E}{>{\raggedright\arraybackslash}p{0.32\textwidth}} 

\newcommand{\opensrcell}[1]{%
    \cellcolor{OpenSourceGreen!#1}{#1}
}

\usepackage{inconsolata}

\usepackage{graphicx}

\title{Can Vision Language Models Be Adaptive in Mathematics Education? A Learner Model-based Rubric Study \includegraphics[scale=0.022]{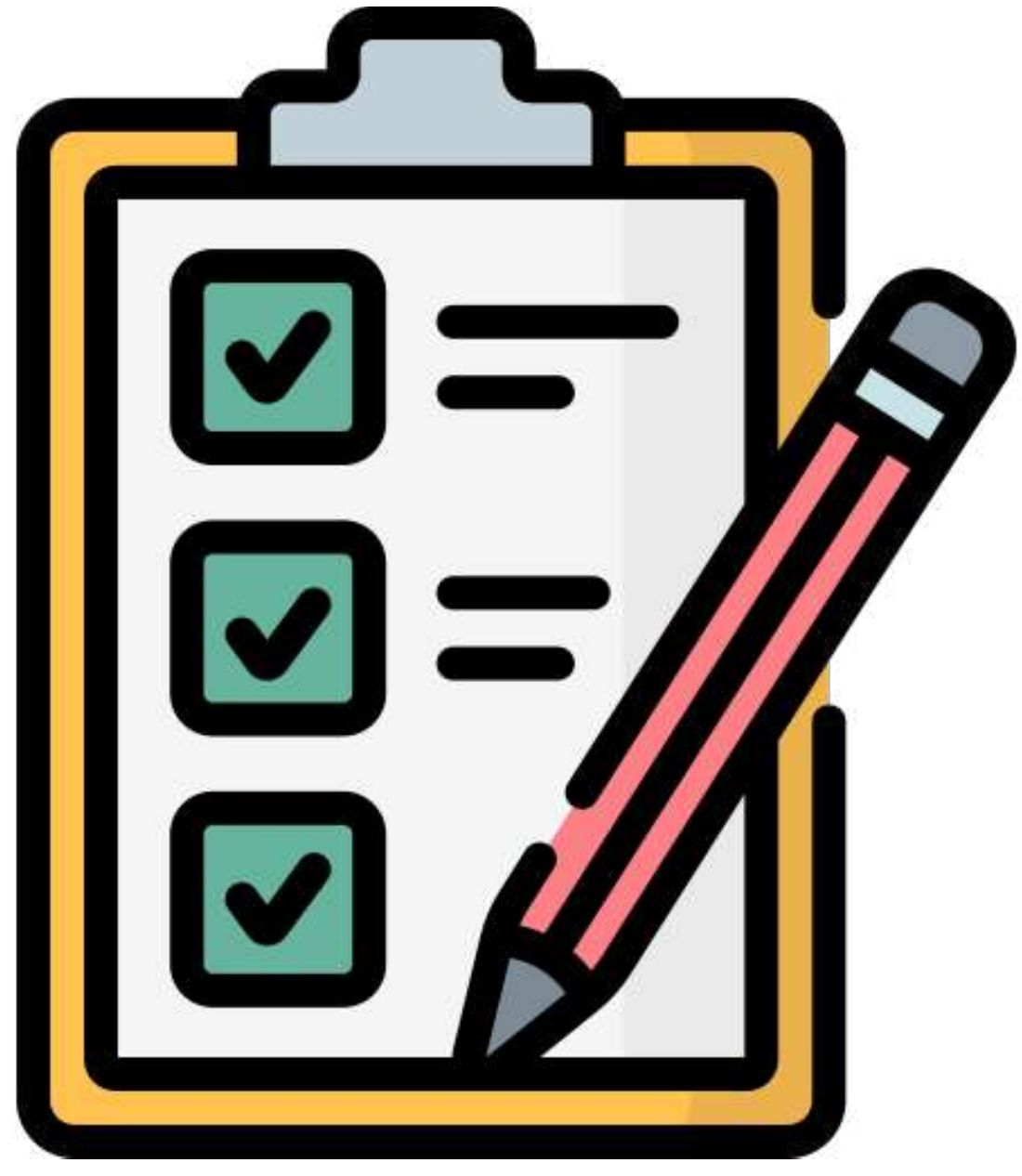}}

\author{
\textbf{Jie Gao}\textsuperscript{1,2}\thanks{Equal contribution.}
\hspace{0.6cm}
\textbf{Yongan Yu}\textsuperscript{1,2}\footnotemark[1]
\hspace{0.6cm}
\textbf{Junzhu Su}\textsuperscript{1}
\\
\textbf{Yiran Lin}\textsuperscript{1}
\hspace{0.6cm}
\textbf{Adam K. Dubé}\textsuperscript{1}\thanks{Corresponding authors.}
\hspace{0.6cm}
\textbf{Jackie Chi Kit Cheung}\textsuperscript{1,2,3}\footnotemark[2]
\\
\textsuperscript{1}McGill University
\hspace{0.5cm}
\textsuperscript{2}Mila -- Quebec AI Institute
\\
\textsuperscript{3}Canada CIFAR AI Chair
\\
{\tt \{jie.gao3, yongan.yu, junzhu.su, yiran.lin\}@mail.mcgill.ca}
\\
{\tt adam.dube@mcgill.ca},
{\tt cheungja@mila.quebec}
}

\begin{document}
\maketitle
\begin{abstract}
Adaptive learning refers to educational technologies that track learners' learning progress and adapt the instructional process based on individual learners' learning performance. It is increasingly recognized as critical for developing an effective learning support tool. Vision language models (VLMs) have seen adoption in mathematics education, and students have been using them as learning aids for personalized instruction. However, it is unknown whether VLMs have the ability to adapt to different learner profiles when providing mathematical instructions.
Current VLMs lack a systematic evaluation framework for this adaptivity to different learner profiles in mathematics tutoring tasks. 
To address this gap, we draw on the learner model from the adaptive learning framework \citep{shute2018adaptive} and propose a learner model-based rubric. Our rubric formalizes adaptivity assessment into three aspects: cognitive aspects, motivational aspects, and complexity. We also evaluate two additional dimensions of VLM responses: correctness (of answers and solutions) and quality (of the response itself). 
Our experimental results show measurable differences in adaptivity across models, and also reveal that current VLMs struggle to consistently produce learner model-based instructional responses, especially when receiving limited learner information.
\end{abstract}

\section{Introduction}
As artificial intelligence shows growing use in education, there is a corresponding rise in large language model-powered tools for providing mathematical instructions and pedagogical support. Learners increasingly adopt VLMs as personal tutoring agents to answer mathematical problems and provide personalized solution feedback \citep{baral2025drawedumath, cherian2024evaluating, gupta2025beyond, liu2024personality}. Personalization is both a defining characteristic of personal tutoring and a critical principle in education, as educators have historically adapted their instruction to provide personalized and differentiated learning support based on diverse learner backgrounds \citep{bernacki2021systematic}.  

\begin{figure}[t!]
    \centering
    \includegraphics[width=\columnwidth]{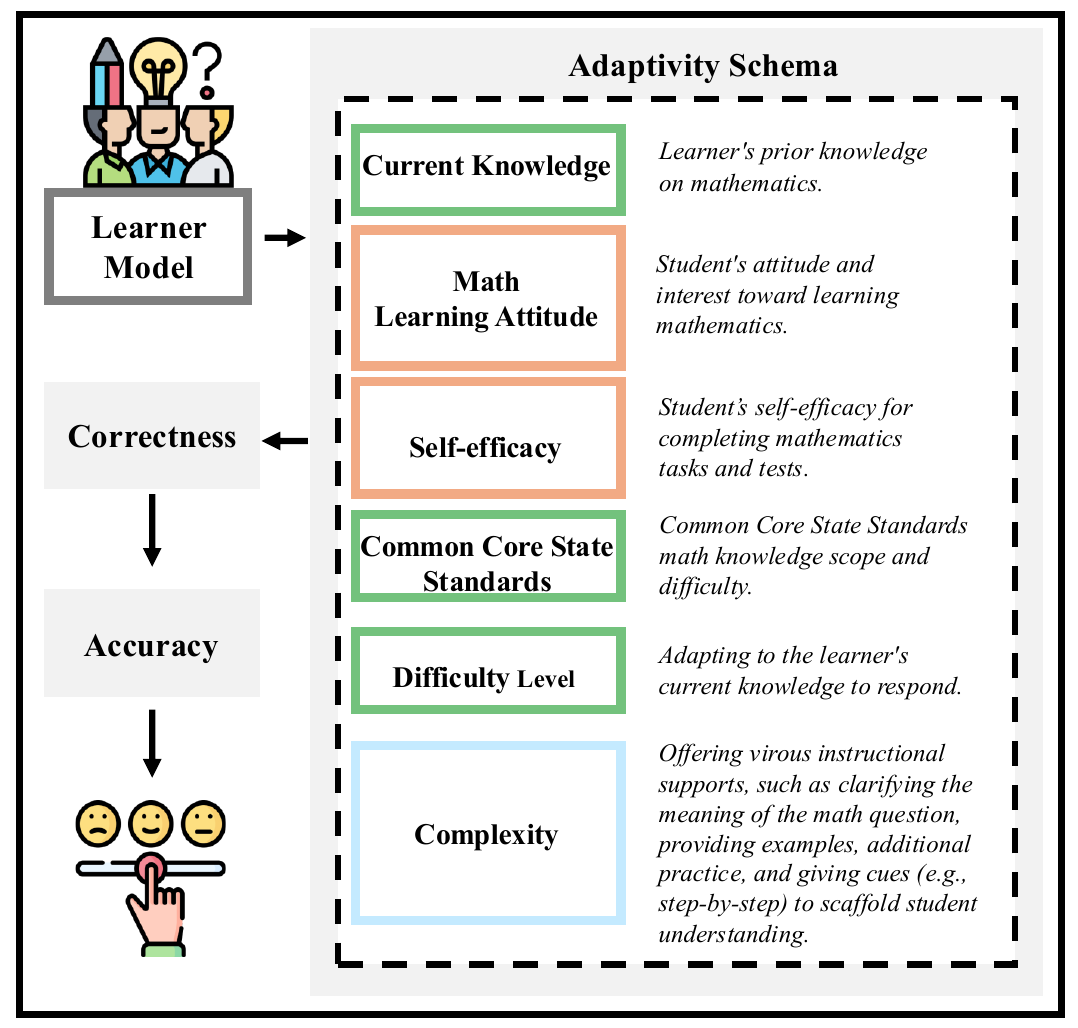}
    \caption{Overview of the learner model-based adaptivity schema. The framework delineates the interplay between learner profile and instructional scaffolding.}
    \vspace{-15pt}
\label{overview}
\end{figure}


To provide personalized learning to address different individuals' learning needs, it is vital to understand learners' learning performance before making instructional decisions. 
Adaptive learning, as one of the most effective educational approaches, involves providing personalized learning instruction and feedback by adapting the learning process to different learners’ needs dynamically \citep{imhof2020implementation, shute2018adaptive, vanzo2025gpt, zheng2022effectiveness}. Its goal is to maximize the effectiveness of instruction through tracking learners’ prior academic background and learning performance, adjusting the content to suit different learners \citep{shute2000individualized}. A foundational component of adaptive learning is the learner model, which captures key aspects of a learner, such as learning performance and learning attitudes, in order to support individualized instruction.  
Several studies have demonstrated that adaptive learning is superior to one-size-fits-all teaching approaches \citep{du2024personalized, wang2023adaptive}. 
Current VLMs primarily collect learner information from conversational interactions \citep{kumar2023impact}. When learners do not explicitly share details about their learning performance, behaviors, or needs, it can be difficult to infer these learners' profiles, which limits instructional decision-making \citep{song2024implementing}. 
Subsequently, it remains unclear whether VLMs generate truly appropriate answers and solutions to support individual learners’ needs \citep{sharma2025role}. 

This paper addresses this critical research gap by developing an adaptive rubric grounded in the learner model of adaptive learning frameworks. 
Our rubric incorporates core evaluation factors from learner models, including cognitive and motivational aspects. 
The rubric aims to assess VLMs' adaptive capabilities in mathematics tutoring tasks and their capacity to deliver pedagogical instruction that aligns with learners' knowledge backgrounds. Accordingly, our research question is: \textbf{"To what extent can VLMs demonstrate adaptive instructional capabilities in visually grounded mathematics tutoring tasks?"}


Our contributions are summarized as follows: (1) We propose a learner model-based rubric to systematically evaluate the adaptivity schema of VLMs in mathematics tutoring tasks, grounded in cognitive, motivational, and complexity dimensions, as illustrated in Figure \ref{overview}. (2) We construct a human-annotated evaluation dataset based on TIMSS mathematics items, incorporating diverse learner profiles and controlled contextual conditions to simulate realistic tutoring scenarios. The constructed model generation pipeline and rubric design are available at our GitHub repository\footnote{GitHub Repo: \url{https://github.com/Michaelyya/Adaptive-LLMs}}. (3) We conduct extensive experiments of five state-of-the-art VLMs, providing first-hand analysis of challenges in generating learner-aware instructional responses, particularly under limited learner information.


\section{Background \& Related Work}


\subsection{Adaptive Learning: The Learner Model}

The learner model forms the foundation of adaptive learning by integrating assessment information and supporting inferences about learners’ proficiencies \citep{shute2018adaptive}. Personalization within adaptive learning systems begins with systematically appraising learner characteristics \citep{zhou2023learning}. 
The systematic review of \citet{bernacki2021systematic} identified key learner characteristics across four dimensions: cognitive factors, beliefs and motivations, social and cultural factors, and emotions and affect. These dimensions directly inform the structural design of our rubric. For instance, understanding a learner’s prior knowledge (a cognitive factor) is essential before providing instruction. Accordingly, our rubric include the current knowledge criterion to evaluate how VLMs identify learners' prior levels to avoid redundant or overly challenging content. 
Similarly, \citet{plass2020toward} categorized variables that can be adapted in instructional responses into four domains: cognitive, motivational, affective, and sociocultural. Within the motivational domain, self-efficacy reflects a learner’s perceived capacity to perform domain-specific tasks. As a result, a learner may display a generally positive attitude toward learning while simultaneously holding low confidence in completing specific mathematics tasks \citep{kaskens2020impact, wen2022systematic}. Our work synthesized insights from previous research and considered the VLMs' conversational attributes for the adaptive rubric construction.


\subsection{The Role of (V)LMs in Education}
In the domain of mathematics, (V)LMs have demonstrated remarkable proficiency as "solvers," achieving superior performance on both text-based (e.g., GSM8K \citep{cobbe2021training}, MathQA \citep{amini2019mathqa}) and vision-based benchmarks (e.g., MATH-Vision \citep{wang2024measuring}, MathVerse \citep{zhang2024mathverse}). However, a critical distinction exists between automated solvers and educational agents: solvers prioritize problem-solving efficiency and final answer correctness \citep{yan-etal-2025-survey}, whereas educational agents must provide pedagogically transparent guidance and scaffolding appropriate to the learner's zone of proximal development \citep{cai2025exploring, yu2025recall}. Recent work suggests that LLMs can mimic tutor-like behavior in interactive sessions \citep{vanzo2025gpt}, and \citet{scarlatos2025exploring} further highlights their potential role in knowledge tracing.
Despite this progress, existing work tends to overlook the multimodal nature of authentic mathematics learning. 
Real-world mathematical tasks often require students to reason across modalities, integrating textual instructions with geometric diagrams, data visualizations, and function plots. 
The emergence of VLMs, including GPT-4V, LLaVA, and Qwen-VL, enables joint visual-textual reasoning \citep{lu2023mathvista}. Current evaluation paradigms, however, largely mirror text-only settings, emphasizing visual question-answering accuracy rather than instructional reasoning quality \citep{lee2025realizing}. Such accuracy-focused evaluations fail to capture ability to explain visual concepts or adapt guidance to a learner’s context, leaving limited empirical evidence on whether multimodal models can serve as adaptive tutors.

\begin{figure*}[t!]
    \centering
    \includegraphics[width=0.9\textwidth]{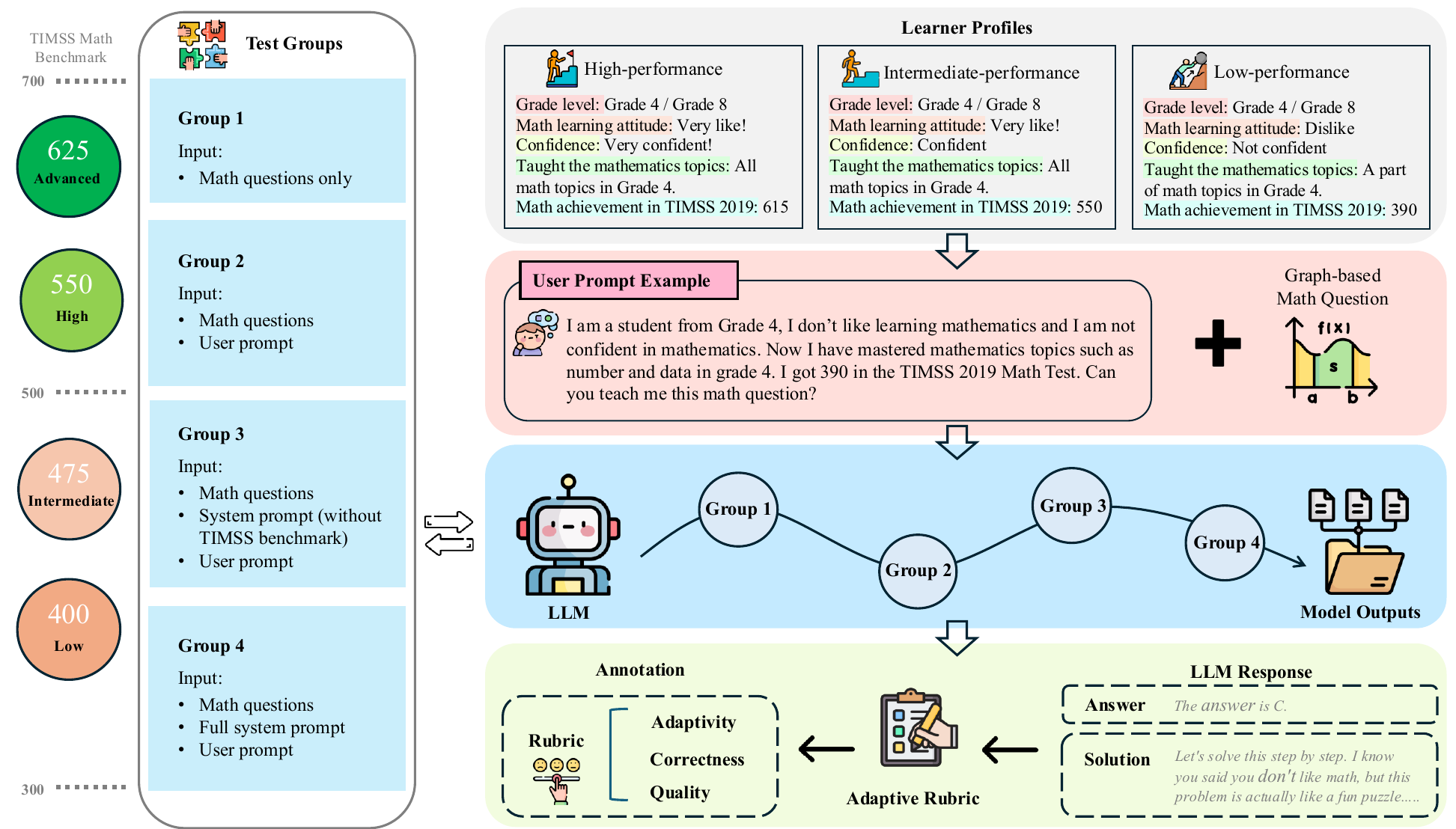}
    \caption{Overview of the adaptive rubric–based evaluation pipeline. Learner profiles and mathematics questions are combined under four test groups with increasing learner-specific information, provided to VLMs to generate instructional responses, and annotated by experts using the proposed rubric.}
\label{full pipeline}
\vspace{-10pt}
\end{figure*}

\subsection{Empirical Study in Adaptive Learning}
The integration of VLMs into educational technologies represents a paradigm shift for intelligent tutoring systems \citep{liu2025advancing}.
While generative model-based systems offer the potential for real-time customization of instructional content based on student behavior \citep{banjade2024empowering, lin2023artificial}, their capacity to deliver contextually appropriate guidance remains questionable. Notably, \citet{borchers2025can} employed a prompt variation framework to assess pedagogical adaptivity, revealing that even state-of-the-art models frequently fail to adjust their feedback strategies when critical context (e.g., student performance or knowledge components) is introduced or removed. 
This suggests a "context insensitivity" where models revert to generic tutoring scripts rather than responding to the specific state of the learner. 
To address this gap, we aim to provide a rigorous, rubric-based assessment of how well VLMs adhere to specific learner profiles.

\section{Rubric Design for Adaptivity Evaluation}




Our rubric is engineered to evaluate whether VLMs demonstrate adaptive instructional intelligence when tailoring support for learners with diverse mathematical proficiencies and backgrounds. From theory to metric, the rubric is structured across three axes: Adaptivity, Correctness, and Quality. This structure was strategically selected to disentangle pedagogical intent from technical accuracy, ensuring that the rubric can isolate a model’s adaptivity from its underlying linguistic and mathematical capabilities.
Among these, Adaptivity constitutes the core evaluative focus, while Correctness and Quality serve as essential baseline controls to ensure mathematical rigor and linguistic clarity. In this work, we operationalize a VLM response into two distinct components: the 'Answer' (the final result) and the 'Solution' (the comprehensive instructional content). The "solution" encompasses all reasoning steps, feedback, and scaffolding strategies beyond the final answer. This proposed rubric assesses whether the model meets the baseline standard of adaptive instructional behavior.
At a high level, the adaptivity dimension assesses whether model responses are appropriately aligned with both the instructional intent of the prompt and the learner's background. The correctness dimension evaluates whether the answers and solutions are mathematically valid. The quality dimension examines overall VLM response quality.  

\textbf{Adaptivity.} 
The adaptivity dimension includes a total of nine key questions (Q1--Q9; see Appendix \ref{appendix: adaptivity}). The evaluation elements of these questions were selected based on the core factors of the adaptive learning environments \citep{bernacki2025handbook, plass2020toward}, while considering the interactive characteristics of VLMs. The primary factors in this rubric include: the learner’s current mathematical knowledge, the learner’s math learning attitude, the learner’s self-efficacy, the solution’s mathematical knowledge scope, the solution difficulty, and the solution complexity. For instance, when a learner presents a negative math learning attitude, an adaptive response should offer positive feedback and instruction to help the learner address the math problem, rather than continuing the conversation in a negative tone or ignoring it \citep{wen2022systematic}. 
In addition, to establish a rigorous and fine-grained normative reference in our rubric,  we integrated Common Core State Standards (CCSS) as a key dimension.
Although TIMSS frameworks provide broad domain categories, CCSS can provide a micro-level taxonomy for evaluation and exhibit 90\% conceptual alignment with TIMSS frameworks\citep{schmidt2012curricular}.
Solution complexity is assessed across Q6--Q9, examining whether the LLM response helps the learner understand and explain the math question, providing relevant examples, offering additional similar math exercises for reinforcement, and delivering step-by-step and detailed instructions.

\textbf{Correctness.} 
This dimension mainly focuses on assessing if the VLM responses provide valid and correct instruction and feedback, addressing the specific math questions. Both the answer and the solution should be correct and consistent to determine the correctness of the response (Q10--Q12; see Appendix \ref{appendix: correctness}). In this dimension, three questions were used to evaluate the correctness of the answer and its solution. Notably, a completely correct solution is defined as one that includes both complete and valid mathematical reasoning and yields a correct final answer. If the solution demonstrates accurate mathematical reasoning but ends with an incorrect final answer, it should be labeled as partially correct.


\textbf{Quality.} 
This dimension focuses primarily on evaluating the overall quality of VLM generation. We conducted fine-grained, multi-dimensional annotations and collected multiple crucial evaluation dimensions from existing benchmarks for summarization quality evaluation \citep{jones2024multi, lee2024unisumeval, tang2024tofueval}, including hallucinations, mathematical reasoning, faithfulness, coherence, consistency, grammatical correctness, formatting adherence, and length control. 
The quality dimension includes a total of seven questions (Q13--Q19; see Appendix \ref{appendix: accuracy}).



\begin{figure*}[t!]
    \centering
    \includegraphics[width=0.9\textwidth]{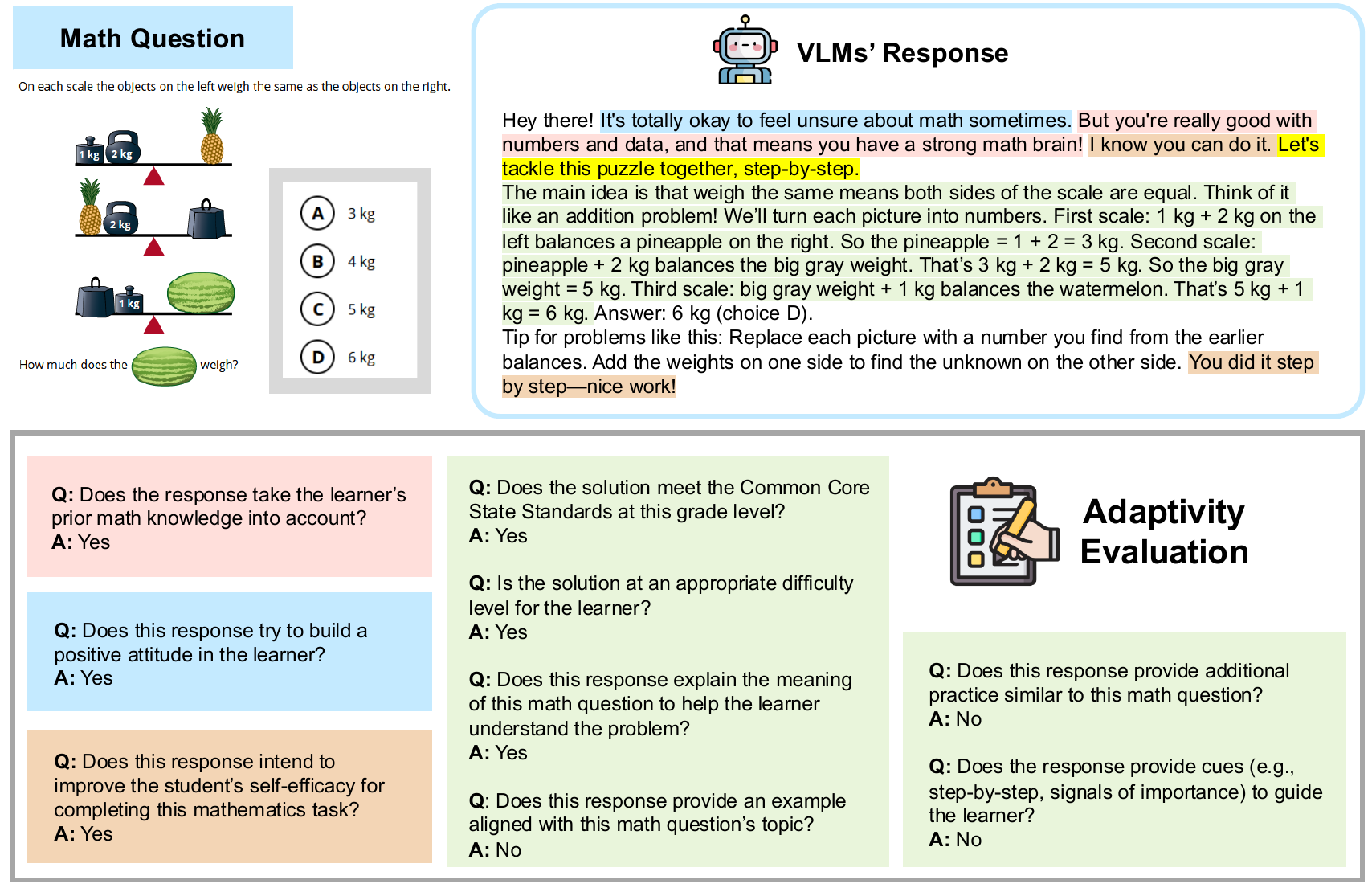}
    \caption{A case study of the adaptive evaluation process for a Grade 4 math problem. The upper panel shows the VLM-generated response with motivational support and step-by-step scaffolding, while the lower panels illustrate the annotation workflow, where Q1--Q9 map directly to the adaptivity rubric in Appendix~\ref{appendix: adaptivity}.}
\label{case study}
\vspace{-10pt}
\end{figure*}

\section{Rubric-Guided Evaluation of VLMs}
To empirically evaluate the adaptive capabilities of VLMs in mathematics tutoring tasks, we implement a structured evaluation pipeline grounded in the proposed learner model–based rubric. As illustrated in Figure~\ref{full pipeline}, the pipeline constructs learner profiles, pairs them with mathematics questions under varying contextual conditions, elicits instructional responses from VLMs, and evaluates the resulting outputs through expert annotation. This methodology adopts a quasi-experimental design that simulates realistic tutoring interactions by systematically varying two key factors: learner profiles and test groups. 
While we acknowledge that such simulated interactions using synthetic learner profiles are not the final answer to evaluating VLMs in the open-ended complexity of real-world classrooms, they represent a necessary and rigorous first step. By leveraging a controlled synthetic environment, we can isolate specific adaptive behaviors and minimize the confounding variables inherent in live discourse between humans and AI. 


\subsection{Dataset and Model Selection}
We draw mathematics items from the Trends in International Mathematics and Science Study (TIMSS) 2019 and 2023 with official permission from the International Association for the Evaluation of Educational Achievement (IEA) \citep{Fishbein2021TIMSS, Fishbein2025TIMSS}. From the full assessment pool, we construct a stratified dataset of $Q=10$ representative questions, with five questions sampled from Grade~4 and five from Grade~8. To emphasize multimodal reasoning, we prioritize visually grounded items, resulting in $90\%$ of the selected questions requiring graphical or diagrammatic interpretation. 
We evaluate a set of $M=5$ VLMs spanning both proprietary and open-source paradigms. The detailed computational costs, model selection and decoding configuration are provided in Appendix \ref{computaional costs} and Appendix \ref{decoding config}.

\subsection{Learner Profiles and Context Conditions}
To simulate realistic tutoring scenarios, we construct learner profiles derived from authentic TIMSS statistics and formalize each learner as a tuple $\ell=(a,c,p,s)$, where $a$ denotes learning attitude, $c$ denotes self-reported confidence, $p$ represents prior topic coverage, and $s$ corresponds to the learner’s TIMSS achievement score; across Grade~4 and Grade~8, we instantiate six profiles spanning three performance tiers (high, intermediate, and low). Building on this formulation, we study context sensitivity by defining four input context conditions $C_k$ with monotonically increasing learner-specific information:
\[
C_1=\{q\}, \quad
C_2=\{q,(a,c,p)\},
\]
\[
C_3=\{q,S,(a,c,p)\}, \quad
C_4=\{q,S,(a,c,p,s)\},
\]

\noindent where $q$ denotes the mathematics question and $S$ denotes a system prompt specifying grade-level learning objectives.

\begin{table*}[t]
\centering
\scriptsize
\setlength{\tabcolsep}{8pt}
\renewcommand{\arraystretch}{1.2}
\begin{tabular}{l|ccc|ccc|ccc}
\toprule
\multirow{2}{*}{\textbf{Model}} & \multicolumn{3}{c|}{\textbf{Adaptivity (Q1-9)}} & \multicolumn{3}{c|}{\textbf{Correctness (Q10-12)}} & \multicolumn{3}{c}{\textbf{Quality (Q13-19)}} \\
\cmidrule(lr){2-4} \cmidrule(lr){5-7} \cmidrule(lr){8-10}
 & \includegraphics[scale=0.045]{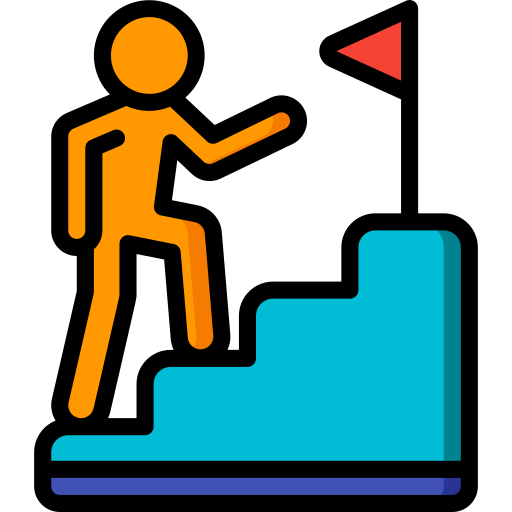} & \includegraphics[scale=0.045]{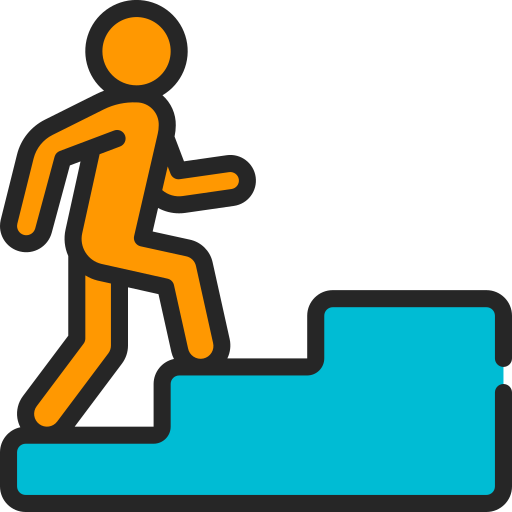} & \includegraphics[scale=0.045]{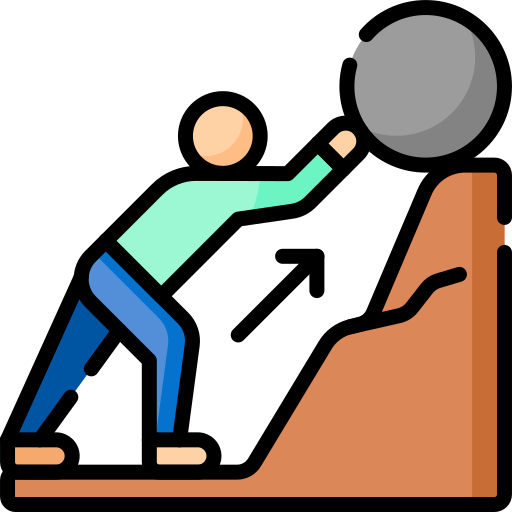}
 & \includegraphics[scale=0.045]{latex/pics/student-high.png} & \includegraphics[scale=0.045]{latex/pics/student-interm.png} & \includegraphics[scale=0.045]{latex/pics/student-low.png}
 & \includegraphics[scale=0.045]{latex/pics/student-high.png} & \includegraphics[scale=0.045]{latex/pics/student-interm.png} & \includegraphics[scale=0.045]{latex/pics/student-low.png} \\
\midrule
GPT-5 \includegraphics[scale=0.02]{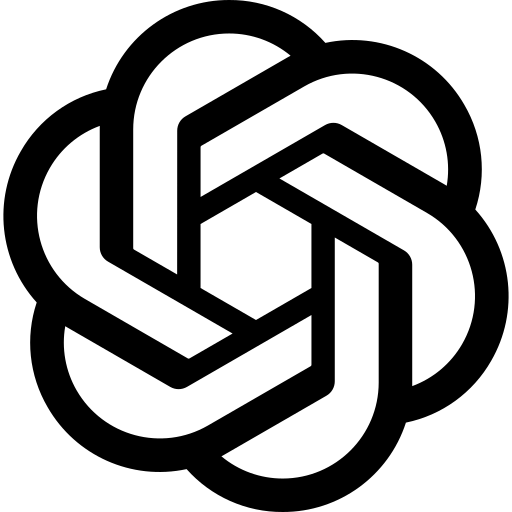} & 52\% & 51\% & \underline{64\%} & \underline{72\%} & \underline{70\%} & \underline{79\%} & \underline{90\%} & \underline{89\%} & \underline{92\%} \\
GPT-o1 \includegraphics[scale=0.02]{latex/pics/gpt-logo.png} & 46\% & 44\% & 47\% & 68\% & 63\% & 65\% & \underline{90\%} & 88\% & 91\% \\
Gemini-2.5-Flash \includegraphics[scale=0.025]{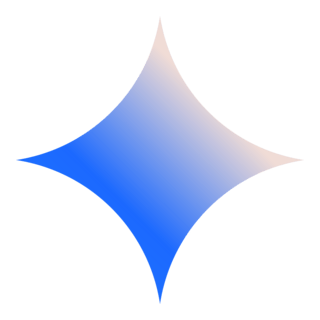} & \underline{65\%} & \underline{68\%} & \textbf{70\%} & \textbf{82\%} & \textbf{80\%} & \textbf{85\%} & \textbf{95\%} & \textbf{94\%} & \textbf{96\%} \\
Llama3.2-11B-VL \includegraphics[scale=0.07]{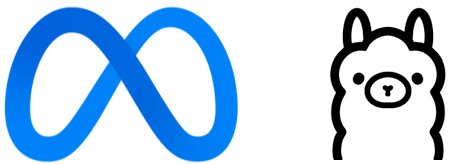} & 50\% & 53\% & 48\% & 19\% & 28\% & 20\% & 65\% & 63\% & 65\% \\
Qwen3-30B-VL \includegraphics[scale=0.004]{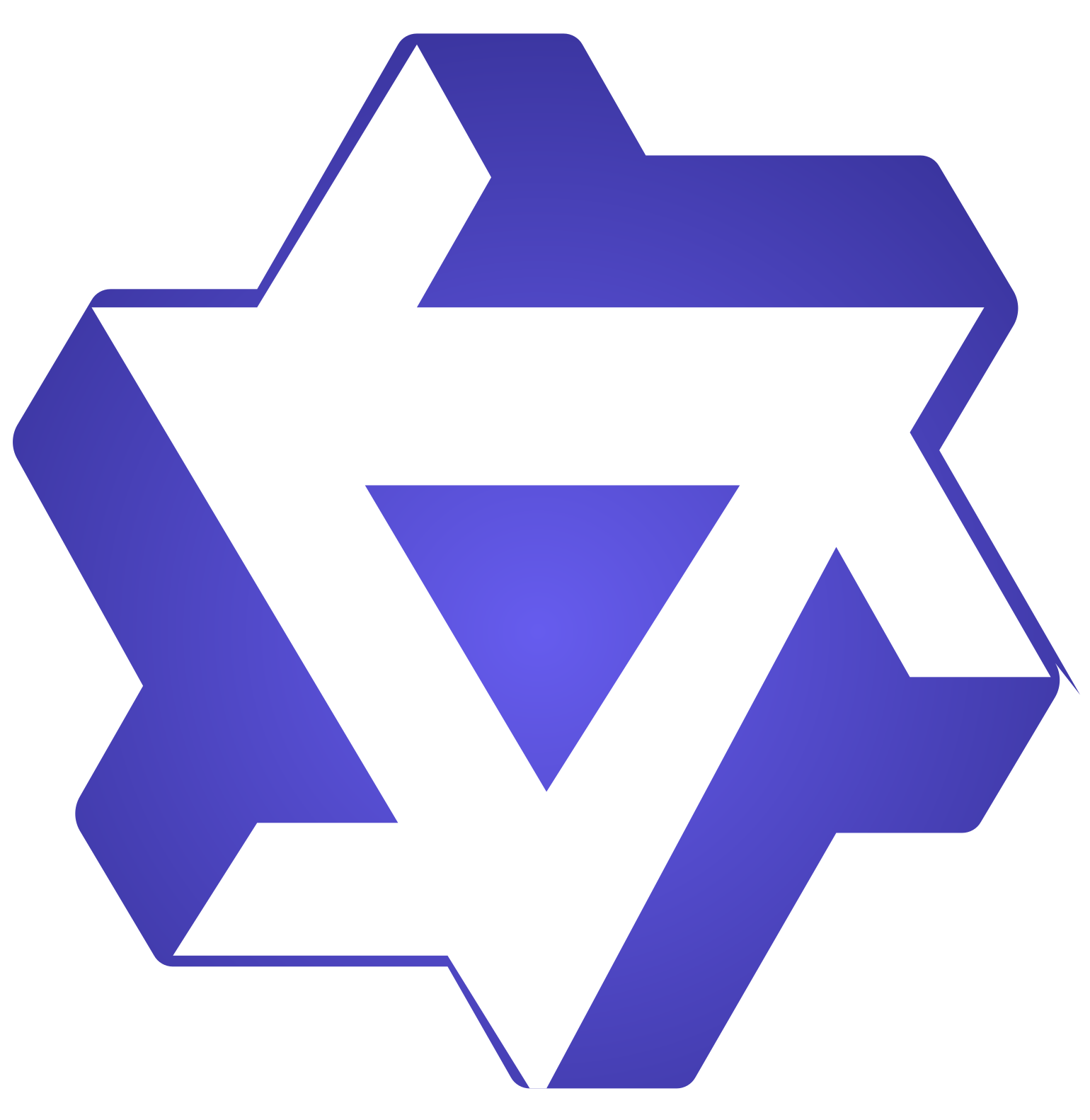} & \textbf{71\%} & \textbf{71\%} & \textbf{70\%} & 56\% & 57\% & 52\% & 76\% & 74\% & 72\% \\
\bottomrule
\end{tabular}
\caption{Performance comparison of VLMs across dimensions. Values represent the mean score percentage achieved across the rubric for each dimension. Learner profiles are denoted by icons: High- (\protect\includegraphics[scale=0.03]{latex/pics/student-high.png}), {Intermediate-} (\protect\includegraphics[scale=0.03]{latex/pics/student-interm.png}), and Low-performance (\protect\includegraphics[scale=0.03]{latex/pics/student-low.png}). \textbf{Bold} denotes the best performance; \underline{underline} denotes the second best.}
\label{tab:overall}
\vspace{-10pt}
\end{table*}

\subsection{Expert Annotation and Evaluation}
Our experimental design follows a full factorial structure. The total volume of model-generated responses is calculated as:
\[
N = \underbrace{|G| \times |P|}_{\text{Learner Profile}} \times \underbrace{|C|}_{\text{Context}} \times \underbrace{|Q|}_{\text{Task}} \times \underbrace{|M|}_{\text{Model}} = 600
\]
where we account for $|G|=2$ grades, $|P|=3$ performance tiers, $|C|=4$ context conditions, $|Q|=5$ questions per grade, and $|M|=5$ models.
Data annotation was conducted by four domain experts trained in the Learning Sciences program. Prior to annotation, each expert completes a one-hour one-on-one training session that introduces the rubric and includes discussion of at least three representative case studies to calibrate the scoring criteria. Each response was evaluated using structured numerical ratings defined by the rubric (see Appendix \ref{sec:detailed_rubric}). 
The rubric items were implemented in a Google Form, and annotators rated each VLM response using this instrument. Adaptivity score and Correctness score were positively keyed, whereas Quality score is derived from error-related items. Specifically, the Quality score is computed as $1 - \frac{1}{7}\sum_{i=13}^{19} e_i$, where $e_i \in \{0,1\}$ indicates whether the corresponding error type (Q13--Q19) is present in the response. All seven error categories are equally weighted in this aggregation, such that higher values consistently indicate better performance. 
Statistical analyses are therefore conducted at the response level rather than the item level. For completeness, we additionally report item-level analysis in Appendix~\ref{appendix:item_analysis}.
A case study is shown in Figure \ref{case study}, the color-coded groupings represent the distinct dimensions of adaptivity. These annotations demonstrate how qualitative text outputs are converted into binary numerical scores (Yes=1, No=0).
To validate the reliability of the annotation process, a subset of 120 responses was independently annotated by core coders, yielding a strong inter-rater agreement with a Cohen's Kappa of $\kappa = 0.87$. Detailed agreement metrics for each model are reported in Appendix~\ref{annotation}.

\begin{table*}[t]
\centering
\scriptsize
\setlength{\tabcolsep}{3pt}
\renewcommand{\arraystretch}{1.6}

\begin{tabular}{ll|*{25}{c}}
\toprule
\multicolumn{2}{c|}{\multirow{4}{*}{\textbf{Dimensions}}} &
\multicolumn{25}{c}{\textbf{Model Performance by Group and Overall (O)}} \\
\cmidrule(lr){3-27}

\multicolumn{2}{c|}{} &
\multicolumn{15}{c|}{\textbf{Proprietary Models} (\%)} &
\multicolumn{10}{c}{\textbf{Open-Source Models} (\%)} \\
\cmidrule(lr){3-17}\cmidrule(lr){18-27}

\multicolumn{2}{c|}{} &
\multicolumn{5}{c}{\textbf{GPT-5} \includegraphics[scale=0.02]{latex/pics/gpt-logo.png}} &
\multicolumn{5}{c}{\textbf{GPT-o1} \includegraphics[scale=0.02]{latex/pics/gpt-logo.png}} &
\multicolumn{5}{c|}{\textbf{Gemini-2.5-flash} \includegraphics[scale=0.025]{latex/pics/gemini-logo.png}} &
\multicolumn{5}{c}{\textbf{Llama3.2-11B-VL} \includegraphics[scale=0.07]{latex/pics/llama-logo.png}} &
\multicolumn{5}{c}{\textbf{Qwen3-30B-VL} \includegraphics[scale=0.004]{latex/pics/Qwen_logo.png}} \\
\cmidrule(lr){3-7}\cmidrule(lr){8-12}\cmidrule(lr){13-17}\cmidrule(lr){18-22}\cmidrule(lr){23-27}

\includegraphics[scale=0.02]{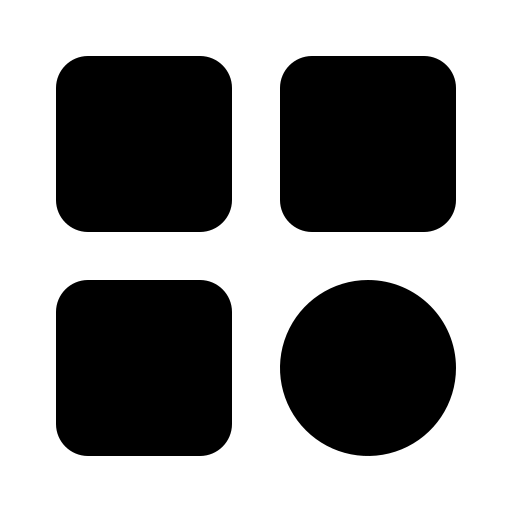} & \textbf{Metric} &
G1 & G2 & G3 & G4 & O &
G1 & G2 & G3 & G4 & O &
G1 & G2 & G3 & G4 & O &
G1 & G2 & G3 & G4 & O &
G1 & G2 & G3 & G4 & O \\
\midrule

\multirow{3}{*}{\rotatebox{90}{\textbf{Cognitive}}}
& CK (Q1)
& \propcell{31} & \propcell{94} & \propcell{94} & \propcell{89} & \propcell{77}
& \propcell{53} & \propcell{66} & \propcell{68} & \propcell{67} & \propcell{63}
& \propcell{53} & \propcell{100} & \propcell{100} & \propcell{100} & \propcell{88}
& \opensrcell{36} & \opensrcell{47} & \opensrcell{83} & \opensrcell{78} & \opensrcell{61}
& \opensrcell{67} & \opensrcell{100} & \opensrcell{100} & \opensrcell{100} & \opensrcell{92} \\

& CCS (Q4)
& \propcell{61} & \propcell{94} & \propcell{94} & \propcell{89} & \propcell{85}
& \propcell{83} & \propcell{97} & \propcell{100} & \propcell{97} & \propcell{94}
& \propcell{83} & \propcell{94} & \propcell{100} & \propcell{100} & \propcell{94}
& \opensrcell{72} & \opensrcell{92} & \opensrcell{72} & \opensrcell{86} & \opensrcell{81}
& \opensrcell{94} & \opensrcell{97} & \opensrcell{94} & \opensrcell{97} & \opensrcell{96} \\

& DL (Q5)
& \propcell{44} & \propcell{94} & \propcell{86} & \propcell{89} & \propcell{78}
& \propcell{61} & \propcell{86} & \propcell{86} & \propcell{89} & \propcell{81}
& \propcell{72} & \propcell{74} & \propcell{89} & \propcell{97} & \propcell{83}
& \opensrcell{67} & \opensrcell{72} & \opensrcell{58} & \opensrcell{75} & \opensrcell{68}
& \opensrcell{92} & \opensrcell{100} & \opensrcell{86} & \opensrcell{86} & \opensrcell{91} \\

\cline{2-27}

\multirow{2}{*}{\rotatebox{90}{\textbf{Motiv.}}}
& LA (Q2)
& \propcell{3} & \propcell{92} & \propcell{89} & \propcell{78} & \propcell{65}
& \propcell{22} & \propcell{43} & \propcell{27} & \propcell{47} & \propcell{35}
& \propcell{6} & \propcell{97} & \propcell{100} & \propcell{100} & \propcell{76}
& \opensrcell{11} & \opensrcell{33} & \opensrcell{86} & \opensrcell{81} & \opensrcell{53}
& \opensrcell{39} & \opensrcell{94} & \opensrcell{100} & \opensrcell{100} & \opensrcell{83} \\

& SE (Q3)
& \propcell{0} & \propcell{53} & \propcell{47} & \propcell{53} & \propcell{38}
& \propcell{8} & \propcell{11} & \propcell{8} & \propcell{3} & \propcell{8}
& \propcell{0} & \propcell{100} & \propcell{97} & \propcell{97} & \propcell{74}
& \opensrcell{3} & \opensrcell{11} & \opensrcell{61} & \opensrcell{53} & \opensrcell{32}
& \opensrcell{17} & \opensrcell{94} & \opensrcell{92} & \opensrcell{97} & \opensrcell{75} \\

\cline{2-27}

\multirow{4}{*}{\rotatebox{90}{\textbf{Complexity}}}
& Expl. (Q6)
& \propcell{6} & \propcell{75} & \propcell{81} & \propcell{61} & \propcell{56}
& \propcell{17} & \propcell{83} & \propcell{51} & \propcell{42} & \propcell{48}
& \propcell{64} & \propcell{100} & \propcell{97} & \propcell{97} & \propcell{90}
& \opensrcell{69} & \opensrcell{86} & \opensrcell{72} & \opensrcell{86} & \opensrcell{78}
& \opensrcell{92} & \opensrcell{100} & \opensrcell{89} & \opensrcell{97} & \opensrcell{94} \\

& Exam. (Q7)
& \propcell{0} & \propcell{8} & \propcell{11} & \propcell{6} & \propcell{6}
& \propcell{6} & \propcell{14} & \propcell{0} & \propcell{0} & \propcell{5}
& \propcell{3} & \propcell{12} & \propcell{8} & \propcell{8} & \propcell{8}
& \opensrcell{3} & \opensrcell{8} & \opensrcell{8} & \opensrcell{3} & \opensrcell{6}
& \opensrcell{8} & \opensrcell{8} & \opensrcell{3} & \opensrcell{0} & \opensrcell{5} \\

& Prac. (Q8)
& \propcell{0} & \propcell{42} & \propcell{44} & \propcell{22} & \propcell{27}
& \propcell{0} & \propcell{0} & \propcell{0} & \propcell{0} & \propcell{0}
& \propcell{0} & \propcell{0} & \propcell{0} & \propcell{0} & \propcell{0}
& \opensrcell{0} & \opensrcell{0} & \opensrcell{8} & \opensrcell{8} & \opensrcell{4}
& \opensrcell{0} & \opensrcell{0} & \opensrcell{0} & \opensrcell{0} & \opensrcell{0} \\

& SBS (Q9)
& \propcell{17} & \propcell{86} & \propcell{86} & \propcell{86} & \propcell{69}
& \propcell{58} & \propcell{86} & \propcell{78} & \propcell{89} & \propcell{78}
& \propcell{92} & \propcell{100} & \propcell{97} & \propcell{94} & \propcell{96}
& \opensrcell{72} & \opensrcell{64} & \opensrcell{58} & \opensrcell{78} & \opensrcell{68}
& \opensrcell{100} & \opensrcell{100} & \opensrcell{100} & \opensrcell{100} & \opensrcell{100} \\

\midrule
\multicolumn{2}{c|}{\textbf{Average}} &
19 & \textbf{68} & \textbf{68} & 63 & 55
& 33 & \textbf{60} & 51 & 52 & 49
& 36 & 77 & \textbf{78} & \textbf{78} & 76
& 37 & 50 & 61 & \textbf{64} & 55
& 45 & \textbf{86} & 85 & \textbf{86} & 81 \\
\bottomrule
\end{tabular}
\caption{Detailed adaptivity dimension performance. Values represent the percentage (\%) of responses satisfying rubric criteria. Row metrics are: Cognitive (CK: Current Knowledge, CCS: Content Scope, DL: Difficulty Level), Motivational (LA: Learning Attitude, SE: Self-Efficacy), and Complexity(Expl.: Explanation, Exam.: Examples, Prac.: Practice, SBS: Step-by-Step). Background colors distinguish \textcolor[rgb]{0.2,0.6,1.0}{\textbf{Proprietary}} vs. \textcolor[rgb]{0.2,0.7,0.2}{\textbf{Open-Source}} models.}
\label{tab: full table}
\vspace{-10pt}
\end{table*}

\section{Results}
We examine the relationship between learner profiles and VLMs' performance in adaptivity, correctness, and quality. Our analysis investigates whether these dimensions exhibit systematic variation across learner profiles and model-related factors (VLMs and test group), reflecting models' sensitivity to learner differences and available contextual information. Table~\ref{tab:overall} summarizes overall performance patterns across VLMs and learner profiles. The reported values represent the mean percentage of total possible points achieved within each dimension. For example, a 70\% Adaptivity score indicates the model satisfied 70\% of the Q1–Q9 criteria. To provide deeper insight into specific adaptive behaviors, Table~\ref{tab: full table} disaggregates the Adaptivity dimension. It reports detailed performance on individual sub-metrics (e.g., Cognitive, Motivational, Complexity) across the four test groups (G1--G4), revealing how the availability of context influences specific instructional strategies.
To provide evidence for these relationships, we employ a series of two-way analyses of variance (ANOVA), with learner profile and model-related factors as independent variables. Further details of the analysis are reported in Appendix~\ref{sec:stat_analysis} and Appendix~\ref{sec:stat_analysis_testgroup}.

\subsection{Experimental analysis}

\textbf{Adaptivity (Q1–Q9).} 
We analyzed how adaptivity scores relate to learner profiles (high-, intermediate-, and low-performing), VLMs, and test groups. Overall, adaptivity scores showed significant relationships across VLMs and across test groups.
Adaptivity varied substantially across models, with a significant main effect of VLMs, $F(4, 705) = 41.81$, $p < .001$, indicating differences in adaptivity performance across models. Tukey HSD post hoc comparisons indicated that Gemini-2.5-Flash and Qwen3-VL-30B achieved significantly higher adaptivity scores than GPT-5, GPT-o1, and Llama3.2-11B-VL, whereas differences between Gemini-2.5-Flash and Qwen3-30B-VL were not statistically significant. In contrast, adaptivity scores showed no significant association with learner profiles, $F(2, 705) = 1.72$, $p = .18$, and no significant interaction between learner profile and VLMs, $F(8, 705) = 1.63$, $p = .11$. This uniformity across groups suggests that these VLMs default to a "one-size-fits-all" pedagogical style. While they could provide answers and solutions, they lack the nuanced scaffolding based on the specific needs of different learner profiles.
A complementary analysis examined adaptivity as a function of learner profile and test group (Groups 1–4, reflecting increasing contextual information). The results showed a significant relationship between contextual information and adaptivity: responses generated with access to additional context exhibited higher adaptivity scores. This pattern was supported by a significant main effect of test group, $F(3, 708) = 92.73$, $p < .001$. Tukey HSD comparisons revealed that Group 1 (math question only) scored significantly lower than Groups 2, 3, and 4. 



\textbf{Correctness (Q10–Q12).} 
Correctness scores showed a similar pattern, with variation driven primarily by model differences, demonstrating a highly significant main effect of VLMs, $F(4, 705) = 64.28$, $p < .001$. Crucially, this model-driven performance was robust and consistent across different learner performance levels. Post hoc analyses indicated that Gemini-2.5-Flash and GPT-based models outperformed Llama3.2-11B-VL, and that Gemini-2.5-Flash also scored significantly higher than Qwen3-30B-VL.
When examining learner profile and VLMs, correctness scores exhibited no association with learner performance level, as reflected by a non-significant main effect of learner profile, $F(2, 705) = 0.01$, $p = .99$, and no significant interaction, $F(8, 705) = 0.65$, $p = .74$. The lack of significant variance suggests two underlying dynamics regarding model behavior: either the baseline user prompt (Group 2) is already sufficient for the VLMs to reach their performance ceiling on this task, or the models ignore the supplementary system prompts provided in Groups 3 and 4.




\textbf{Quality (Q13–Q19).} 
Quality scores further reinforced the pattern observed for adaptivity and correctness: learner profiles did not systematically relate to the model response quality, whereas substantial differences emerged across VLMs.
However, quality differed significantly across models, as indicated by a significant main effect of VLMs, $F(4, 705) = 69.44$, $p < .001$. Tukey HSD comparisons showed that Llama3.2-11B-VL produced significantly higher error counts than all other models, while Gemini-2.5-Flash achieved significantly lower error scores than GPT-5, GPT-O1, and Qwen3-30B-VL.



Consistent with the overall analyses, distinct performance profiles emerged across the models. Gemini-2.5-Flash demonstrated strong and balanced performance across all metrics, while GPT-o1 prioritized correctness and quality over adaptivity.
Qwen3-30B-VL obtained the highest average adaptivity score across learner profiles (M = $71\%$). However, its correctness and quality were comparatively lower. This is mainly due to overly long model generations that were more likely to be truncated or incomplete, which in turn produced formatting issues and partial non-adherence to prompt instructions. 
We also observed a correctness gap between Llama3.2-11B-VL and the closed-source models, with the latter consistently outperforming Llama3.2-11B-VL. 


\subsection{Observations}
This subsection summarizes key empirical observations derived from rubric-based annotations of VLM responses. 
These observations are grounded in aggregated rubric scores across learner profiles and context conditions, supplemented by qualitative inspection of representative model outputs. 


\paragraph{VLMs exhibit minimal behavioral adaptation to learner attitudes and self-efficacy.}
\noindent As adaptivity constitutes the core dimension of the proposed rubric, we first examine how VLMs respond to learner-specific motivational factors, particularly learning attitude and self-efficacy (Q3–Q5). Across models, GPT-5, GPT-o1, and Llama3.2-11B-Vision-Instruct exhibit consistently low response rates in addressing these factors. For example, GPT-o1 addresses self-efficacy concerns in only $8\%$ of applicable cases, with a $0\%$ response rate for Grade~4 intermediate-performing learners. In contrast, low-performance learners receive substantially more motivational feedback, with an average response rate of $36.8\%$ when low confidence is explicitly indicated. Among all evaluated models, Qwen3-30B-VL ($33.1\%$) and Gemini-2.5-Flash ($32.5\%$) demonstrate comparatively stronger engagement with learners’ self-efficacy. Interestingly, some Qwen3-30B-VL responses include explicit encouragement (e.g., \texttt{“You got this!”}), suggesting partial instructional responses using motivational language. These patterns suggest models implicitly equate lower performance with greater motivational need, overlooking the importance of supporting confidence and learning attitudes among high- and intermediate-performing learners.


\paragraph{Providing learners' knowledge backgrounds to the VLM does not guarantee appropriate feedback.}
\noindent It is essential to understand learners' current knowledge level before providing them with feedback. However, such an understanding does not mean that the VLMs have the capabilities to provide appropriate differentiated feedback. In several cases, responses to different learners are nearly identical, suggesting weak sensitivity to learner distinctions. The VLMs frequently misunderstand how to deliver suitable content to specific groups. Table \ref{tab: full table} illustrates that GPT-5 generates longer and more detailed explanations for high-performance learners than for intermediate-performance learners, despite the latter typically requiring greater instructional scaffolding. These findings indicate that access to learner background information alone does not guarantee appropriate or pedagogically aligned feedback.

\paragraph{Multimodal misinterpretation and response quality failures.}

\noindent Beyond adaptivity, our rubric-based evaluation surfaces notable correctness and quality issues, particularly for visually grounded mathematics problems. Geometry items pose a significant challenge: for instance, a Grade 4 geometry question (G4Q5) required comparing four choices’ rectangles to identify the correct one. Although the task structure is clear, only 10 out of 60 model responses produced correct answers. This error pattern persists despite the problem’s clear structure, pointing to limitations in visual–spatial reasoning rather than ambiguity in the task itself. These errors are consistently reflected in low correctness scores across multiple models and learner profiles. 
Moreover, a common failure mode involves conflating multiple-choice option labels with their embedded numerical content.
When presented with options like "D. 6kg", the correct response should identify the label (D) rather than the embedded value (6kg). However, models regarded the embedded value as the final answer, omitting the required option label. 
Additionally, when questions integrate images, formulas, and text, models often extract misaligned or inaccurate information. This manifests as hallucinations or inaccurate extraction in solution, compromising mathematical reasoning and instructional reliability, even when the final answer appears superficially plausible.

\section{Conclusion}
This paper evaluates whether VLMs demonstrate adaptive capabilities in mathematics tutoring tasks. To this end, we introduce a learner model–inspired rubric that provides a structured framework for assessing adaptivity beyond surface-level personalization.
Applying this rubric to VLMs across learner profiles reveals that, while models exhibit some sensitivity to learner context, their adaptive behavior remains limited and inconsistent. In particular, models often fail to differentiate instructional strategies across learners with distinct academic backgrounds, indicating a gap between generic tutoring responses and genuinely adaptive instruction.

The proposed rubric offers a principled tool for systematically comparing adaptive capabilities across models. Looking forward, future work may extend this rubric to distinguish baseline adaptivity from more sophisticated forms of instructional adaptation and to examine additional components of adaptive learning frameworks. More broadly, this work highlights the need for evaluation methodologies that prioritize pedagogical alignment with learners, providing guidance for the development of more adaptive and educationally grounded multimodal models.

\section*{Limitations}
While we have developed the adaptive rubric and accompanying worksheet to identify the VLMs' adaptive capabilities, the rubric questions for adaptivity are not exhaustive. Other adaptivity variables, including those from affective and socio-cultural aspects, are also important for evaluation. What's more, the limited math items used in this study may make our evaluation suffer from insufficient item coverage regarding pedagogical adaptivity. In addition, our study simulated the learner rather than testing with real students, which cannot include all the learning behavior factors of real students. Moreover, there might be other differences inherent in using an automated tutoring system compared to human tutors that are not addressed in the current adaptive learning framework. For instance, human tutors may create more social pressure, which might be helpful or harmful for learning. A further limitation is that the proposed rubric is designed as a baseline that evaluates whether responses have adaptive capabilities in mathematics tutoring tasks. However, a response classified as adaptive in this study does not represent a high-quality, high-level adaptive response.

\section*{Ethical Consideration}
This study involves the evaluation of vision language models using synthetic and publicly available mathematical problem data. No personally identifiable information or human subject data were used in model evaluation. We obtained official permission from the International Association for the Evaluation of Educational Achievement (IEA) to use the Trends in International Mathematics and Science Study (TIMSS) 2019 and 2023 mathematics items from Grade 4 and Grade 8. Thus, the data has no potential privacy or content safety concerns. Additionally, the learner characteristics of learner profile data were extracted from the TIMSS 2019 and TIMSS 2023 public reports and datasets \citep{Fishbein2021TIMSS, Fishbein2025TIMSS}.

\bibliography{custom}

@manual{Fishbein2025TIMSS,
  title = {TIMSS 2023 User Guide for International Database},
  author = {Fishbein, B. and Taneva, M. and Kowolik, K.},
  year = {2025},
  institution = {Boston College, TIMSS \& PIRLS International Study Center},
  url = {https://timss2023.org/data},
  note = {Retrieved from TIMSS 2023 International Database}
}

@article{sharma2025role,
  title={The role of large language models in personalized learning: a systematic review of educational impact},
  author={Sharma, Sahil and Mittal, Puneet and Kumar, Mukesh and Bhardwaj, Vivek},
  journal={Discover Sustainability},
  volume={6},
  number={1},
  pages={1--24},
  year={2025},
  publisher={Springer}
}

@article{kumar2023impact,
  title={Impact of guidance and interaction strategies for LLM use on Learner Performance and perception},
  author={Kumar, Harsh and Musabirov, Ilya and Reza, Mohi and Shi, Jiakai and Wang, Xinyuan and Williams, Joseph Jay and Kuzminykh, Anastasia and Liut, Michael},
  journal={arXiv preprint arXiv:2310.13712},
  year={2023}
}

@manual{Fishbein2021TIMSS,
  title = {TIMSS 2019 User Guide for the International Database},
  author = {Fishbein, B. and Foy, P. and Yin, L.},
  year = {2021},
  edition = {2nd},
  organization = {Boston College, TIMSS \& PIRLS International Study Center},
  url = {https://timssandpirls.bc.edu/timss2019/international-database/}
}

@inproceedings{borchers2025can,
  title={Can large language models match tutoring system adaptivity? a benchmarking study},
  author={Borchers, Conrad and Shou, Tianze},
  booktitle={International Conference on Artificial Intelligence in Education},
  pages={407--420},
  year={2025},
  organization={Springer}
}

@article{lee2025realizing,
  title={Realizing visual question answering for education: GPT-4V as a multimodal AI},
  author={Lee, Gyeonggeon and Zhai, Xiaoming},
  journal={TechTrends},
  volume={69},
  number={2},
  pages={271--287},
  year={2025},
  publisher={Springer}
}

@article{lu2023mathvista,
  title={Mathvista: Evaluating mathematical reasoning of foundation models in visual contexts},
  author={Lu, Pan and Bansal, Hritik and Xia, Tony and Liu, Jiacheng and Li, Chunyuan and Hajishirzi, Hannaneh and Cheng, Hao and Chang, Kai-Wei and Galley, Michel and Gao, Jianfeng},
  journal={arXiv preprint arXiv:2310.02255},
  year={2023}
}

@inproceedings{vanzo2025gpt,
  title={GPT-4 as a homework tutor can improve student engagement and learning outcomes},
  author={Vanzo, Alessandro and Chowdhury, Sankalan Pal and Sachan, Mrinmaya},
  booktitle={Proceedings of the 63rd Annual Meeting of the Association for Computational Linguistics (Volume 1: Long Papers)},
  pages={31119--31136},
  year={2025}
}

@inproceedings{yu2025recall,
  title={From Recall to Reasoning: Automated Question Generation for Deeper Math Learning Through Large Language Models},
  author={Yu, Yongan and Krantz, Alexandre and Lobczowski, Nikki G},
  booktitle={International Conference on Artificial Intelligence in Education},
  pages={414--422},
  year={2025},
  organization={Springer}
}

@article{wang2024measuring,
  title={Measuring multimodal mathematical reasoning with math-vision dataset},
  author={Wang, Ke and Pan, Junting and Shi, Weikang and Lu, Zimu and Ren, Houxing and Zhou, Aojun and Zhan, Mingjie and Li, Hongsheng},
  journal={Advances in Neural Information Processing Systems},
  volume={37},
  pages={95095--95169},
  year={2024}
}

@article{cai2025exploring,
  title={Exploring the impact of integrating AI tools in higher education using the Zone of Proximal Development},
  author={Cai, Lianyu and Msafiri, Mgambi Msambwa and Kangwa, Daniel},
  journal={Education and Information Technologies},
  volume={30},
  number={6},
  pages={7191--7264},
  year={2025},
  publisher={Springer}
}

@inproceedings{zhang2024mathverse,
  title={Mathverse: Does your multi-modal llm truly see the diagrams in visual math problems?},
  author={Zhang, Renrui and Jiang, Dongzhi and Zhang, Yichi and Lin, Haokun and Guo, Ziyu and Qiu, Pengshuo and Zhou, Aojun and Lu, Pan and Chang, Kai-Wei and Qiao, Yu and others},
  booktitle={European Conference on Computer Vision},
  pages={169--186},
  year={2024},
  organization={Springer}
}

@inproceedings{amini2019mathqa,
  title={Mathqa: Towards interpretable math word problem solving with operation-based formalisms},
  author={Amini, Aida and Gabriel, Saadia and Lin, Shanchuan and Koncel-Kedziorski, Rik and Choi, Yejin and Hajishirzi, Hannaneh},
  booktitle={Proceedings of the 2019 conference of the North American chapter of the association for computational linguistics: Human language technologies, volume 1 (long and short papers)},
  pages={2357--2367},
  year={2019}
}

@article{cobbe2021training,
  title={Training verifiers to solve math word problems},
  author={Cobbe, Karl and Kosaraju, Vineet and Bavarian, Mohammad and Chen, Mark and Jun, Heewoo and Kaiser, Lukasz and Plappert, Matthias and Tworek, Jerry and Hilton, Jacob and Nakano, Reiichiro and others},
  journal={arXiv preprint arXiv:2110.14168},
  year={2021}
}

@article{banjade2024empowering,
  title={Empowering education by developing and evaluating generative AI-Powered tutoring system for enhanced student learning},
  author={Banjade, Shivraj and Patel, Hiran and Pokhrel, Sangita and others},
  journal={Journal of Artificial Intelligence and Capsule Networks},
  volume={6},
  number={3},
  pages={278--298},
  year={2024},
  publisher={IRO Journals}
}

@article{lin2023artificial,
  title={Artificial intelligence in intelligent tutoring systems toward sustainable education: a systematic review},
  author={Lin, Chien-Chang and Huang, Anna YQ and Lu, Owen HT},
  journal={Smart learning environments},
  volume={10},
  number={1},
  pages={41},
  year={2023},
  publisher={Springer}
}

@article{liu2025advancing,
  title={Advancing education through tutoring systems: A systematic literature review},
  author={Liu, Vincent and Latif, Ehsan and Zhai, Xiaoming},
  journal={arXiv preprint arXiv:2503.09748},
  year={2025}
}

@inproceedings{scarlatos2025exploring,
  title={Exploring knowledge tracing in tutor-student dialogues using llms},
  author={Scarlatos, Alexander and Baker, Ryan S and Lan, Andrew},
  booktitle={Proceedings of the 15th International Learning Analytics and Knowledge Conference},
  pages={249--259},
  year={2025}
}

@inproceedings{yan-etal-2025-survey,
    title = "A Survey of Mathematical Reasoning in the Era of Multimodal Large Language Model: Benchmark, Method {\&} Challenges",
    author = "Yan, Yibo  and
      Su, Jiamin  and
      He, Jianxiang  and
      Fu, Fangteng  and
      Zheng, Xu  and
      Lyu, Yuanhuiyi  and
      Wang, Kun  and
      Wang, Shen  and
      Wen, Qingsong  and
      Hu, Xuming",
    editor = "Che, Wanxiang  and
      Nabende, Joyce  and
      Shutova, Ekaterina  and
      Pilehvar, Mohammad Taher",
    booktitle = "Findings of the Association for Computational Linguistics: ACL 2025",
    month = jul,
    year = "2025",
    address = "Vienna, Austria",
    publisher = "Association for Computational Linguistics",
    url = "https://aclanthology.org/2025.findings-acl.614/",
    doi = "10.18653/v1/2025.findings-acl.614",
    pages = "11798--11827",
    ISBN = "979-8-89176-256-5"
}

@article{plass2020toward,
  title={Toward a taxonomy of adaptivity for learning},
  author={Plass, Jan L and Pawar, Shashank},
  journal={Journal of Research on Technology in Education},
  volume={52},
  number={3},
  pages={275--300},
  year={2020},
  publisher={Taylor \& Francis}
}

@book{bernacki2025handbook,
  title={Handbook of Personalized Learning},
  author={Bernacki, Matthew L and Walkington, Candace and Emery, Alyssa and Zhang, Ling},
  year={2025},
  publisher={Taylor \& Francis}
}

@article{imhof2020implementation,
  title={Implementation of adaptive learning systems: Current state and potential},
  author={Imhof, Christof and Bergamin, Per and McGarrity, St{\'e}phanie},
  journal={Online teaching and learning in higher education},
  pages={93--115},
  year={2020},
  publisher={Springer}
}

@incollection{shute2018adaptive,
  title={Adaptive e-learning},
  author={Shute, Valerie and Towle, Brendon},
  booktitle={Aptitude},
  pages={105--114},
  year={2018},
  publisher={Routledge}
}

@article{shute2000individualized,
  title={Individualized and group approaches to training},
  author={Shute, VJ and Lajoie, SP and Gluck, KA},
  journal={Training and retraining: A handbook for business, industry, government, and the military},
  pages={171--207},
  year={2000}
}

@article{bernacki2021systematic,
  title={A systematic review of research on personalized learning: Personalized by whom, to what, how, and for what purpose (s)?},
  author={Bernacki, Matthew L and Greene, Meghan J and Lobczowski, Nikki G},
  journal={Educational Psychology Review},
  volume={33},
  number={4},
  pages={1675--1715},
  year={2021},
  publisher={Springer}
}

@article{kaskens2020impact,
  title={Impact of Children’s math self-concept, math self-efficacy, math anxiety, and teacher competencies on math development},
  author={Kaskens, Jarise and Segers, Eliane and Goei, Sui Lin and Van Luit, Johannes Eh and Verhoeven, Ludo},
  journal={Teaching and teacher education},
  volume={94},
  pages={103096},
  year={2020},
  publisher={Elsevier}
}

@article{wen2022systematic,
  title={A systematic review of secondary students’ attitudes towards mathematics and its relations with mathematics achievement},
  author={Wen, Run and Dub{\'e}, Adam K},
  journal={Journal of Numerical Cognition},
  volume={8},
  number={2},
  pages={295--325},
  year={2022}
}

@article{wang2023adaptive,
  title={When adaptive learning is effective learning: comparison of an adaptive learning system to teacher-led instruction},
  author={Wang, Shuai and Christensen, Claire and Cui, Wei and Tong, Richard and Yarnall, Louise and Shear, Linda and Feng, Mingyu},
  journal={Interactive learning environments},
  volume={31},
  number={2},
  pages={793--803},
  year={2023},
  publisher={Taylor \& Francis}
}

@inproceedings{gupta2025beyond,
  title={Beyond final answers: Evaluating large language models for math tutoring},
  author={Gupta, Adit and Reddig, Jennifer and Calo, Tommaso and Weitekamp, Daniel and MacLellan, Christopher J},
  booktitle={International Conference on Artificial Intelligence in Education},
  pages={323--337},
  year={2025},
  organization={Springer}
}

@article{baral2025drawedumath,
  title={DrawEduMath: Evaluating Vision Language Models with Expert-Annotated Students' Hand-Drawn Math Images},
  author={Baral, Sami and Lucy, Li and Knight, Ryan and Ng, Alice and Soldaini, Luca and Heffernan, Neil T and Lo, Kyle},
  journal={arXiv preprint arXiv:2501.14877},
  year={2025}
}

@article{cherian2024evaluating,
  title={Evaluating large vision-and-language models on children's mathematical olympiads},
  author={Cherian, Anoop and Peng, Kuan-Chuan and Lohit, Suhas and Matthiesen, Joanna and Smith, Kevin and Tenenbaum, Josh},
  journal={Advances in Neural Information Processing Systems},
  volume={37},
  pages={15779--15800},
  year={2024}
}

@article{zheng2022effectiveness,
  title={The effectiveness of technology-facilitated personalized learning on learning achievements and learning perceptions: A meta-analysis},
  author={Zheng, Lanqin and Long, Miaolang and Zhong, Lu and Gyasi, Juliana Fosua},
  journal={Education and Information Technologies},
  volume={27},
  number={8},
  pages={11807--11830},
  year={2022},
  publisher={Springer}
}

@article{marraffino2021adapting,
  title={Adapting training in real time: an empirical test of adaptive difficulty schedules},
  author={Marraffino, Matthew D and Schroeder, Bradford L and Fraulini, Nicholas W and Van Buskirk, Wendi L and Johnson, Cheryl I},
  journal={Military Psychology},
  volume={33},
  number={3},
  pages={136--151},
  year={2021},
  publisher={Taylor \& Francis}
}

@inproceedings{zhou2023learning,
  title={Learning to predict persona information for dialogue personalization without explicit persona description},
  author={Zhou, Wangchunshu and Li, Qifei and Li, Chenle},
  booktitle={Findings of the Association for Computational Linguistics: ACL 2023},
  pages={2979--2991},
  year={2023}
}

@misc{GPT-5,
url={https://cdn.openai.com/gpt-5-system-card.pdf}, journal={GPT-5 system card openai August 13, 2025 1}}

@article{hackett1989exploration,
  title={An exploration of the mathematics self-efficacy/mathematics performance correspondence},
  author={Hackett, Gail and Betz, Nancy E},
  journal={Journal for research in Mathematics Education},
  volume={20},
  number={3},
  pages={261--273},
  year={1989},
  publisher={National Council of Teachers of Mathematics}
}

@inproceedings{liu2024personality,
  title={Personality-aware student simulation for conversational intelligent tutoring systems},
  author={Liu, Zhengyuan and Yin, Stella Xin and Lin, Geyu and Chen, Nancy},
  booktitle={Proceedings of the 2024 Conference on Empirical Methods in Natural Language Processing},
  pages={626--642},
  year={2024}
}

@article{kibrislioglu2015investigation,
  title={An investigation about 6th grade students’ attitudes towards mathematics},
  author={Kibrislioglu, Nermin},
  journal={Procedia-Social and Behavioral Sciences},
  volume={186},
  pages={64--69},
  year={2015},
  publisher={Elsevier}
}

@inproceedings{lee2024unisumeval,
  title={UniSumEval: Towards unified, fine-grained, multi-dimensional summarization evaluation for LLMs},
  author={Lee, Yuho and Yun, Taewon and Cai, Jason and Su, Hang and Song, Hwanjun},
  booktitle={Findings of the Association for Computational Linguistics: EMNLP 2024},
  pages={3941--3960},
  year={2024}
}

@inproceedings{jones2024multi,
  title={A multi-aspect framework for counter narrative evaluation using large language models},
  author={Jones, Jaylen and Mo, Lingbo and Fosler-Lussier, Eric and Sun, Huan},
  booktitle={Proceedings of the 2024 Conference of the North American Chapter of the Association for Computational Linguistics: Human Language Technologies (Volume 2: Short Papers)},
  pages={147--168},
  year={2024}
}

@inproceedings{tang2024tofueval,
  title={TofuEval: Evaluating hallucinations of LLMs on topic-focused dialogue summarization},
  author={Tang, Liyan and Shalyminov, Igor and Wong, Amy and Burnsky, Jon and Vincent, Jake and Yang, Yu’an and Singh, Siffi and Feng, Song and Song, Hwanjun and Su, Hang and others},
  booktitle={Proceedings of the 2024 Conference of the North American Chapter of the Association for Computational Linguistics: Human Language Technologies (Volume 1: Long Papers)},
  pages={4455--4480},
  year={2024}
}

@article{schmidt2012curricular,
  title={Curricular coherence and the common core state standards for mathematics},
  author={Schmidt, William H and Houang, Richard T},
  journal={Educational Researcher},
  volume={41},
  number={8},
  pages={294--308},
  year={2012},
  publisher={Sage Publications Sage CA: Los Angeles, CA}
}

@article{du2024personalized,
  title={Personalized adaptive learning in higher education: A scoping review of key characteristics and impact on academic performance and engagement},
  author={Du Plooy, Eileen and Casteleijn, Daleen and Franzsen, Denise},
  journal={Heliyon},
  volume={10},
  number={21},
  year={2024},
  publisher={Elsevier}
}

@article{song2024implementing,
  title={Implementing the dynamic feedback-driven learning optimization framework: a machine learning approach to personalize educational pathways},
  author={Song, Chuanxiang and Shin, Seong-Yoon and Shin, Kwang-Seong},
  journal={Applied Sciences},
  volume={14},
  number={2},
  pages={916},
  year={2024},
  publisher={MDPI}
}


\appendix

\section{Detailed Rubric}
\label{sec:detailed_rubric}

This section comprehensively documents the rubric criteria, scoring hierarchies, and reference examples established in the evaluation framework. To ensure reproducibility and transparency, each entry below specifies the exact text used to prompt evaluators, the theoretical dimensions underpinning the metric (e.g., \textit{Cognitive Aspects} vs. \textit{Motivational Aspects}), and the precise logic required to assign values (e.g., $-1$ vs. $0$ vs. $1$). Additionally, we provide the canonical examples that define the boundary conditions for each score, presenting the material verbatim from the study's design documents.

\subsection{Adaptivity}
\label{appendix: adaptivity}
\begin{rubricitem}{Q1. Does the response take the learner’s prior math knowledge into account?}{Cognitive aspects - Current knowledge \citep{plass2020toward}}\\
\textbf{Criteria:} The response should match the learner’s stated math level. A response fails when it: 1) The content area is more than the learner's. 2) The content area is out of this grade's scope. \\
    \textbf{Scoring \& Examples:}
    \begin{itemize}
        \item \textbf{0 = No}: Prompt: I just learned algebra in grade 8. Could you teach me this question "$y = ab$. Which is the value of y when a = 1 and b = 2" \newline Response: Sure! Think about a x b in a higher-dimensional space...
        \item \textbf{1 = Yes}
    \end{itemize}
\end{rubricitem}

\begin{rubricitem}{Q2. Does this response try to build a positive attitude in the learner?}{Motivational aspects - Learning attitude \citep{bernacki2021systematic}}\\
    \textbf{Criteria:} Definition of Learning Attitude in Math: \citet{kibrislioglu2015investigation} defines attitude towards mathematics as liking or disliking of the subject; a tendency to engage in or avoid mathematical activities; a belief that one is good or bad at mathematics; and a belief that mathematics is useful or useless. \\
    \textbf{Scoring \& Examples:}
    \begin{itemize}
        \item \textbf{-1 = No, it responds negatively}
        \item \textbf{0 = No}
        \item \textbf{1 = Yes}: Prompt: I don't like math. Can you help me understand 2 × 3? \newline Response: "I can help you with that...."
    \end{itemize}
\end{rubricitem}

\begin{rubricitem}{Q3. Does this response intend to improve the student’s self-efficacy for completing this mathematics task?}{Motivational aspects - Self-efficacy \citep{plass2020toward}}\\
    \textbf{Criteria:} Math self-efficacy is a situational or problem-specific assessment of an individual's confidence in her or his ability to successfully perform or accomplish a particular mathematical task or problem \citep{hackett1989exploration}. The response should help the student feel capable of solving the math question. \\
    \textbf{Scoring \& Examples:}
    \begin{itemize}
        \item \textbf{-1 = No, it responds negatively}
        \item \textbf{0 = No}
        \item \textbf{1 = Yes}: Prompt: I am not confident in math learning. Can you help me understand 2 × 3? \newline Response: "Math can be tricky, but we can solve this together."
    \end{itemize}
\end{rubricitem}

\begin{rubricitem}{Q4. Does the solution meet the Common Core State Standards at this grade level?}{Cognitive aspects - Difficulty \citep{marraffino2021adapting}}\\
    \textbf{Criteria:} The response aligns with the Common Core standard math knowledge scope and difficulty. \\
    \textbf{Scoring \& Examples:}
    \begin{itemize}
        \item \textbf{0 = No}: Not followed the Common Core Standard
        \item \textbf{1 = Yes}: Prompt: I’m not sure I can solve 12 × 13. Can you teach me? \newline Response: Of course! This is a Number and Operation question in Grade 4......?
    \end{itemize}
\end{rubricitem}

\begin{rubricitem}{Q5. Is the solution at an appropriate difficulty level for the learner? }{Cognitive aspects - Difficulty \citep{marraffino2021adapting}}\\
    \textbf{Criteria:} The response should match the learner’s skill level. A response fails when the explanation is too advanced, too complex, or uses concepts the learner is not ready for. \\
    \textbf{Scoring \& Examples:}
    \begin{itemize}
        \item \textbf{0 = No}: Not provided a solution at an appropriate difficulty level
        \item \textbf{1 = Yes}: Prompt: What is $2 + 2$? \newline Response: It is 4 because adding two items to another two items makes four items.
    \end{itemize}
\end{rubricitem}

\begin{rubricitem}{Q6. Does this response explain the meaning of this math question to help the learner understand the problem?}{Complexity} \\
\textbf{Criteria:} The response should explain the meaning of the math question to help the learner know its logic and better know how to solve it. A response fails when the response does not provide an explanation of the meaning of the math question. \\
    \textbf{Scoring \& Examples:}
    \begin{itemize}
        \item \textbf{0 = No}: Not explained
        \item \textbf{1 = Yes}: Response: ...This question means...?
    \end{itemize}
\end{rubricitem}

\begin{rubricitem}{Q7. Does this response provide an example aligned with this math question’s topic?}{Complexity}\\
    \textbf{Criteria:} The response should give an example that matches the math topic in the question. A response fails when the example is unrelated, off-topic, or does not help illustrate the math idea being asked about. \\
    \textbf{Scoring \& Examples:}
    \begin{itemize}
        \item \textbf{0 = No}: No example provided
        \item \textbf{1 = Yes}: Response: ...Here is an example...
    \end{itemize}
\end{rubricitem}

\begin{rubricitem}{Q8. Does this response provide additional practice similar to this math question?}{Complexity}\\
    \textbf{Criteria:} The response should give extra practice problems that match the math concept in the question. A response fails when the response does not provide extra practice. \\
    \textbf{Scoring \& Examples:}
    \begin{itemize}
        \item \textbf{0 = No}: No additional practice provided
        \item \textbf{1 = Yes}: Response: ...Let's practice...
    \end{itemize}
\end{rubricitem}

\begin{rubricitem}{Q9. Does the response provide cues (e.g., step-by-step, signals of importance) to guide the learner?}{Complexity}\\
    \textbf{Criteria:} The response should include helpful cues such as steps, hints, or pointing out important parts of the problem. A response fails when it gives the answer with no guidance or offers no structure to help the learner follow the process. \\
    \textbf{Scoring \& Examples:}
    \begin{itemize}
        \item \textbf{0 = No}: No cues provided
        \item \textbf{1 = Yes}: Response: ...Let’s go step by step. First...
    \end{itemize}
\end{rubricitem}

\subsection{Correctness}
\label{appendix: correctness}
\begin{rubricitem}{Q10. Does the response provide a correct answer to this math question?}{The response provides a correct answer.}\\
    \textbf{Criteria:} The final answer provided by the model must be mathematically correct. \\
    \textbf{Scoring \& Examples:}
    \begin{itemize}
        \item \textbf{0 = No}: Prompt: $2 + 2 = ?$ \newline Response: 5
        \item \textbf{1 = Yes}: Prompt: $2 + 2 = ?$ \newline Response: 4
    \end{itemize}
\end{rubricitem}

\begin{rubricitem}{Q11. Does the solution support and align with the answer?}

    \textbf{Criteria:} The key explanation and logic should clearly match the final answer.  \\
    \textbf{Scoring \& Examples:}
    \begin{itemize}
        \item \textbf{1 = No}
        \item \textbf{0 = Yes}: Prompt: Can you teach me how to solve 12 × 13?\newline Response: The answer is 156. This is a number \& operation math question...
    \end{itemize}
\end{rubricitem}

\begin{rubricitem}{Q12. Does the response provide a correct solution to this math question?}

\textbf{Criteria:} The response should give the correct and accurate solution based on the math problem.  \\  
    \textbf{Scoring \& Examples:}
    \begin{itemize}
        \item \textbf{0 = No} : More than 50\% incorrect
        \item \textbf{1 = Mostly correct}: Minor error that doesn't affect logic
        \item \textbf{2 = Completely correct}
    \end{itemize}
\end{rubricitem}

\subsection{Quality}
\label{appendix: accuracy}
\begin{rubricitem}{Q13. Does the response have hallucinations or inaccurately extracted information?}
\textbf{Criteria:} The model responds with content that contains hallucinations and/or extracts inaccurate information from the provided prompt. \\
    \textbf{Scoring \& Examples:}
    \begin{itemize}
        \item \textbf{1 = No}
        \item \textbf{0 = Yes}
    \end{itemize}
\end{rubricitem}

\begin{rubricitem}{Q14. Does the response have failures in mathematical or logical reasoning?} \textbf{Criteria:} The model in its responses makes errors in arithmetic calculations, statistical analysis, or other quantitative conceptual or numerical operations.  \\
    \textbf{Scoring \& Examples:}
    \begin{itemize}
        \item \textbf{1 = No}
        \item \textbf{0 = Yes}
    \end{itemize}
\end{rubricitem}

\begin{rubricitem}{Q15. Does the response fail to accurately and completely follow the prompt instructions?}
\textbf{Criteria:} The model doesn't follow explicit or clearly implied instructions in the model input. This includes responding in the wrong language, ignoring the math exercise's key information, or not adhering to the user's requirement. \\
    \textbf{Scoring \& Examples:}
    \begin{itemize}
        \item \textbf{1 = No}
        \item \textbf{0 = Yes}
    \end{itemize}
\end{rubricitem}

\begin{rubricitem}{Q16. Is the response incoherent?}
\textbf{Criteria:} The model's reasoning contains logical fallacies, contradicts itself within the same response, or reaches conclusions that don't follow from the premises. It includes cases where the model presents information and then contradicts it later in the same response. \\
    \textbf{Scoring \& Examples:}
    \begin{itemize}
        \item \textbf{1 = No}
        \item \textbf{0 = Yes}
    \end{itemize}
\end{rubricitem}

\begin{rubricitem}{Q17. Does the response provide irrelevant information?} \textbf{Criteria:} The model in its responses includes information that doesn't address the user's query or goes beyond the scope of what was asked. This includes discussing tangential topics, adding unnecessary background, or focusing on aspects not central to the query. \\
    \textbf{Scoring \& Examples:}
    \begin{itemize}
        \item \textbf{1 = No}
        \item \textbf{0 = Yes}
    \end{itemize}
\end{rubricitem}

\begin{rubricitem}{Q18. Does the response have formatting, grammatical, or vocabulary errors?}
    \textbf{Criteria:} This failure mode occurs when the model's response contains errors in formatting, grammar, or word usage. \\
    \textbf{Scoring \& Examples:}
    \begin{itemize}
        \item \textbf{1 = No}
        \item \textbf{0 = Yes}
    \end{itemize}
\end{rubricitem}

\begin{rubricitem}{Q19. Is the response overly brief or lacking necessary detail?}
    \textbf{Criteria:} The model provides a response that lacks sufficient detail or explanation to adequately address the query, given the complexity of the topic and the apparent depth of information requested. \\
    \textbf{Scoring \& Examples:}
    \begin{itemize}
        \item \textbf{1 = No}
        \item \textbf{0 = Yes}
    \end{itemize}
\end{rubricitem}

\section{Model Sources and Computational Cost}
\label{computaional costs}
The models evaluated in this paper are obtained from the following sources:

\begin{enumerate}
    \item \textbf{GPT-5} and \textbf{GPT-o1} are provided by OpenAI. The corresponding API documentation is available at \url{https://platform.openai.com/docs/models}.
    \item \textbf{Gemini-2.5-flash} is provided by Google Gemini, with API documentation available at \url{https://ai.google.dev/gemini-api/docs}.
    \item \textbf{Qwen3-VL-30B-A3B-Instruct}\footnote{\url{https://huggingface.co/Qwen/Qwen3-VL-30B-A3B-Instruct}} and \textbf{Llama-3.2-11B-Vision-Instruct}\footnote{\url{https://huggingface.co/meta-llama/Llama-3.2-11B-Vision-Instruct}} are open-source base model weights obtained from Hugging Face (\url{https://huggingface.co/}).
\end{enumerate}

For large proprietary models (e.g., GPT-5), a one-time evaluation on 600 samples incurs a cost of approximately \$6 for generating explanations. All open-source model evaluations are conducted on a system equipped with two NVIDIA RTX 4090 GPUs (32GB memory each). Overall, the modest computational requirements demonstrate that our evaluation protocol is accessible to researchers with limited computational resources, while still enabling comprehensive assessment of state-of-the-art models.

\section{Decoding configuration}
\label{decoding config}
To ensure comparability across models, we standardize the decoding parameters for all evaluated VLMs. Specifically, we set the maximum generation length to 512 tokens and use a temperature of 0.7 for all models. Generation terminates when the model reaches the token limit or produces an end-of-sequence token.

These settings are chosen to allow sufficient space for step-by-step explanations while preventing excessively long outputs. The token limit is determined based on pilot annotations by human experts, which estimate the typical length required to provide a complete pedagogical explanation for the evaluated mathematics problems. Importantly, the same configuration is applied across all models to ensure that differences in response quality reflect model behavior rather than decoding differences.

\section{Item-Level Robustness Analysis}
\label{appendix:item_analysis}

To examine whether specific evaluation dimensions drive the reported findings, we conduct an item-level robustness analysis across the 19 evaluation questions used in our human annotation framework. Each item captures a distinct aspect of instructional quality, including learner adaptation, motivational support, instructional scaffolding, correctness, and potential failure modes.

For each item, we compute the mean percentage across the six learner profiles for each model. Table~\ref{tab:item_level_results} reports the resulting item-level averages.
\begin{table}[t]
\small
\begin{tabular}{lccccc}
\toprule
\textbf{Item} & \textbf{Gemini} & \textbf{GPT-5} & \textbf{GPT-O1} & \textbf{Llama} & \textbf{Qwen} \\
\midrule
\multicolumn{6}{c}{\textit{Instructional Adaptivity and Pedagogical Quality}} \\
\midrule
Q1 & 88.0 & 76.5 & 63.5 & 64.2 & 92.2 \\
Q2 & 75.7 & 63.5 & 32.2 & 53.0 & 82.3 \\
Q3 & 73.3 & 36.0 & 7.7 & 32.8 & 74.3 \\
Q4 & 94.2 & 83.7 & 93.7 & 78.7 & 95.3 \\
Q5 & 83.3 & 78.7 & 80.3 & 67.7 & 90.5 \\
Q6 & 90.5 & 53.8 & 48.5 & 77.2 & 94.5 \\
Q7 & 8.5 & 7.2 & 5.7 & 6.3 & 5.7 \\
Q8 & 0.0 & 26.0 & 0.0 & 4.3 & 0.0 \\
Q9 & 95.0 & 66.2 & 75.7 & 67.3 & 100.0 \\
Q10 & 79.2 & 71.5 & 54.3 & 13.5 & 45.0 \\
Q11 & 83.8 & 78.0 & 76.5 & 48.3 & 76.7 \\
\midrule
\multicolumn{6}{c}{\textit{Failure Modes (Lower is Better)}} \\
\midrule
Q13 & 5.3 & 10.5 & 16.8 & 48.8 & 22.3 \\
Q14 & 11.5 & 19.0 & 23.3 & 64.0 & 30.8 \\
Q15 & 7.0 & 17.0 & 22.8 & 59.2 & 43.5 \\
Q16 & 0.7 & 0.8 & 1.7 & 20.0 & 4.3 \\
Q17 & 4.8 & 5.8 & 2.5 & 12.3 & 17.0 \\
Q18 & 2.3 & 0.5 & 0.0 & 13.7 & 43.2 \\
Q19 & 6.3 & 17.0 & 6.3 & 20.3 & 22.5 \\
\bottomrule
\end{tabular}
\caption{Item-level mean percentages across learner profiles (N=6).}
\label{tab:item_level_results}
\end{table}
The results reveal consistent patterns across items. Gemini and Qwen achieve the highest scores on most positive instructional dimensions, including prior-knowledge adaptation (Q1), motivational support (Q2–Q3), explanation quality (Q6), and instructional guidance (Q9). In contrast, GPT-O1 and Llama exhibit substantially higher failure rates in reasoning errors (Q14), instruction-following failures (Q15), and incoherence (Q16).

Importantly, these patterns remain stable across evaluation items, suggesting that the reported model differences are not driven by a small subset of evaluation criteria. Instead, the relative strengths and weaknesses of models appear consistently across multiple pedagogical dimensions.

\section{Human Annotation Details}
\label{annotation}

To verify the reliability of our evaluation rubric, we conducted a human annotation study on a stratified subset of 120 model responses. Two expert annotators independently evaluated these instances, achieving an overall inter-rater agreement of 87.11\%. As detailed in Table~\ref{tab:model_agreement}, these results demonstrate strong consistency across different models and learner profiles, confirming the reproducibility of our scoring framework.

\begin{table}[H] 
\centering
\begin{tabular}{lc}
\toprule
\textbf{Model} & \textbf{Agreement (\%)} \\
\midrule
GPT-o1 & 100.00 \\
GPT-5 & 89.47 \\
Qwen3-VL-30B & 84.21 \\
Gemini-2.5-Flash & 73.68 \\
Llama3.2-11B-VL & 73.68 \\
\midrule
\textbf{Average} & \textbf{87.11} \\
\bottomrule
\end{tabular}
\caption{Detailed inter-rater agreement metrics for each evaluated model on the annotated subset ($N=120$). The overall average agreement across all models is 87.11\%.}
\label{tab:model_agreement} 
\end{table}

\onecolumn

\section{Statistical Analysis of Model Performance}
\label{sec:stat_analysis}

\[
p(a_j, b_j \mid D_{t+1})
\;\propto\;
\underbrace{
p(R_{\text{new}} \mid a_j, b_j, \theta_{\text{new}})
}_{\substack{
\text{Likelihood: how well new responses} \\
\text{fit the current item parameters}
}}
\;\cdot\;
\underbrace{
p(a_j, b_j \mid D_t)
}_{\substack{
\text{Prior:  learned from} \\
\text{previous calibration data}
}}
\]

We conducted a Repeated Measures ANOVA to statistically evaluate the performance differences between VLMs and test items. As detailed in Table \ref{tab:anova_results}, the analysis revealed a statistically significant main effect for the LLM ($F = 41.39, p < .001$), confirming that the choice of model significantly impacts performance quality.

Additionally, we observed a significant main effect for Item ($F = 721.61, p < .001$) and a significant Item $\times$ LLM Interaction ($F = 21.75, p < .001$). This interaction indicates that the performance gap between models is not uniform; rather, it varies depending on the specific difficulty or type of math question presented.\\

\begin{table}[h]
\centering
\small
\renewcommand{\arraystretch}{1.2}
\begin{tabular}{l c c c c c}
\toprule
\textbf{Cases} & \textbf{Sum of Squares} & \textbf{df} & \textbf{Mean Square} & \textbf{F} & \textbf{p} \\
\midrule
\multicolumn{6}{l}{\textit{\textbf{Within Subjects Effects}}} \\
Item & 593.60\textsuperscript{a} & 8\textsuperscript{a} & 74.201\textsuperscript{a} & 721.61\textsuperscript{a} & $< .001$ \\
Item $*$ LLM & 71.56\textsuperscript{a} & 32\textsuperscript{a} & 2.236\textsuperscript{a} & 21.75\textsuperscript{a} & $< .001$ \\
Residuals & 588.17 & 5720 & 0.103 & & \\
\multicolumn{6}{l}{\footnotesize \textsuperscript{a} \textit{Mauchly's test indicates sphericity assumption is violated ($p<.05$).}} \\

\midrule
\multicolumn{6}{l}{\textit{\textbf{Between Subjects Effects}}} \\
LLM & 61.25 & 4 & 15.312 & 41.39 & $< .001$ \\
Residuals & 264.49 & 715 & 0.370 & & \\
\bottomrule
\end{tabular}
\caption{Repeated Measures ANOVA Results. Analysis of Within-Subjects Effects (Item, Interaction) and Between-Subjects Effects (LLM).}
\label{tab:anova_results}
\end{table}

\begin{figure}[h]
    \centering
    \includegraphics[width=0.85\textwidth]{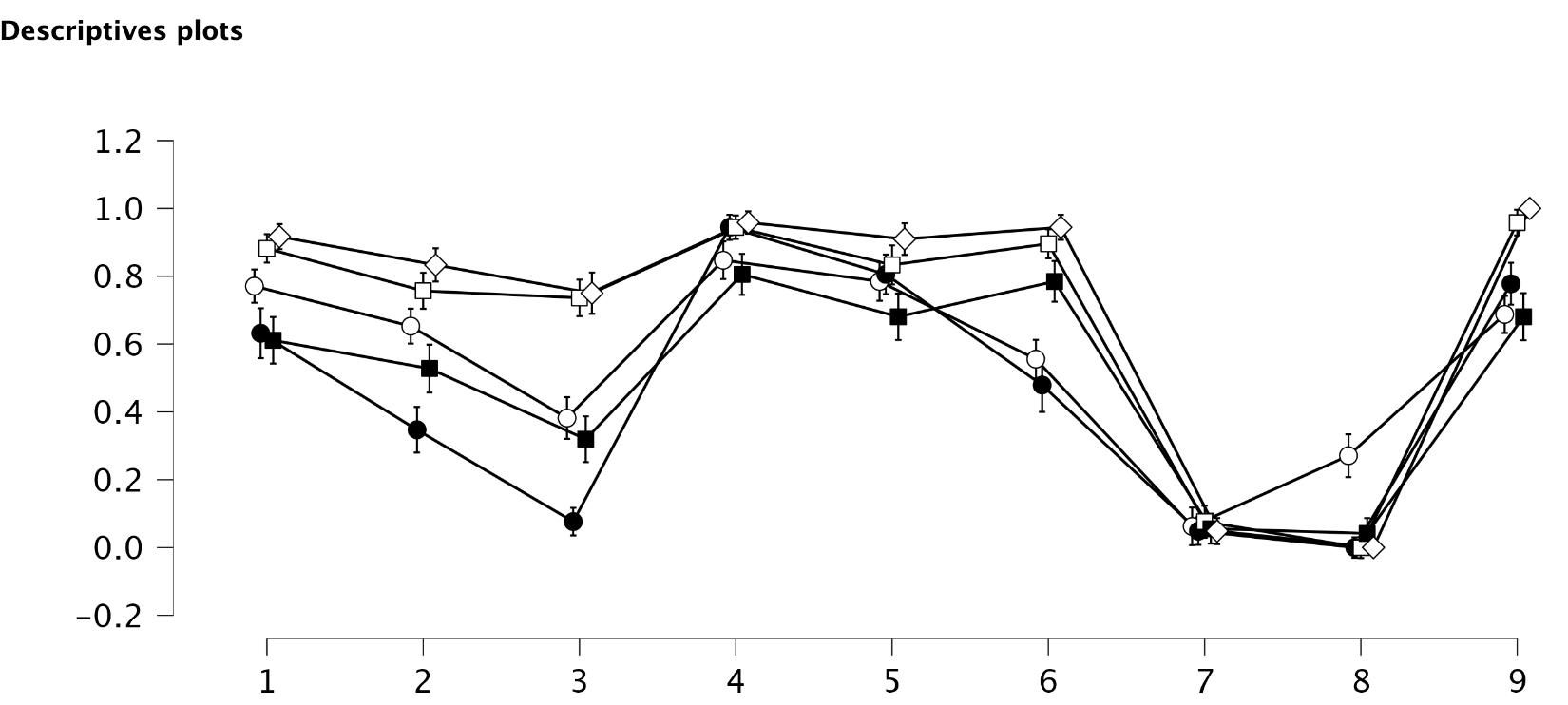} 
    \caption{Descriptive Plots of Model Performance by Item. The x-axis represents the math items, and the y-axis represents the normalized performance score. The divergence of lines illustrates the Item $\times$ LLM interaction effect.}
    \label{fig:descriptives_plot}
\end{figure}

\noindent Figure \ref{fig:descriptives_plot} visualizes these descriptive trends. While some items (e.g., Item 7) proved uniformly difficult for all models, causing performance to converge near zero, other items (e.g., Item 1 and 9) exhibit distinct stratification, where higher-performing models separate clearly from lower-performing ones.

\section{Statistical Analysis by Test Group}
\label{sec:stat_analysis_testgroup}

We extended our analysis to evaluate performance differences between distinct Test Groups using a Repeated Measures ANOVA. As presented in Table \ref{tab:anova_testgroup}, we found a statistically significant main effect for TestGroup ($F(3, 716) = 92.73, p < .001$), with a substantial effect size ($\eta_p^2 = 0.280$). This indicates that the grouping factor accounts for 28\% of the variance in between-subjects performance.

Furthermore, the Item $\times$ TestGroup interaction was significant ($F(24, 5728) = 19.13, p < .001$), suggesting that performance gaps between groups are dependent on the specific item being solved.

\begin{table*}[t] 
\centering
\small
\renewcommand{\arraystretch}{1.2}
\setlength{\tabcolsep}{5pt}

\begin{tabular}{l c c c c c c c}
\toprule
\textbf{Cases} & \textbf{SS} & \textbf{df} & \textbf{MS} & \textbf{F} & \textbf{p} & $\bm{\eta_p^2}$ & $\bm{\omega^2}$ \\
\midrule
\multicolumn{8}{l}{\textit{\textbf{Within Subjects Effects}}} \\
Item & 593.64\textsuperscript{a} & 8\textsuperscript{a} & 74.205\textsuperscript{a} & 695.91\textsuperscript{a} & $< .001$ & 0.493 & 0.412 \\
Item $*$ TestGroup & 48.95\textsuperscript{a} & 24\textsuperscript{a} & 2.040\textsuperscript{a} & 19.13\textsuperscript{a} & $< .001$ & 0.074 & 0.022 \\
Residuals & 610.78 & 5728 & 0.107 & & & & \\
\multicolumn{8}{l}{\footnotesize \textsuperscript{a} \textit{Mauchly's test indicates sphericity assumption is violated ($p<.05$).}} \\

\midrule
\multicolumn{8}{l}{\textit{\textbf{Between Subjects Effects}}} \\
TestGroup & 91.14 & 3 & 30.381 & 92.73 & $< .001$ & 0.280 & 0.088 \\
Residuals & 234.59 & 716 & 0.328 & & & & \\
\bottomrule
\end{tabular}
\caption{Repeated Measures ANOVA (Test Group). Comparison of Within-Subjects and Between-Subjects effects, including effect sizes ($\eta_p^2$ and $\omega^2$).}
\label{tab:anova_testgroup}
\end{table*}

\begin{figure}[h]
    \centering
    \includegraphics[width=0.85\textwidth]{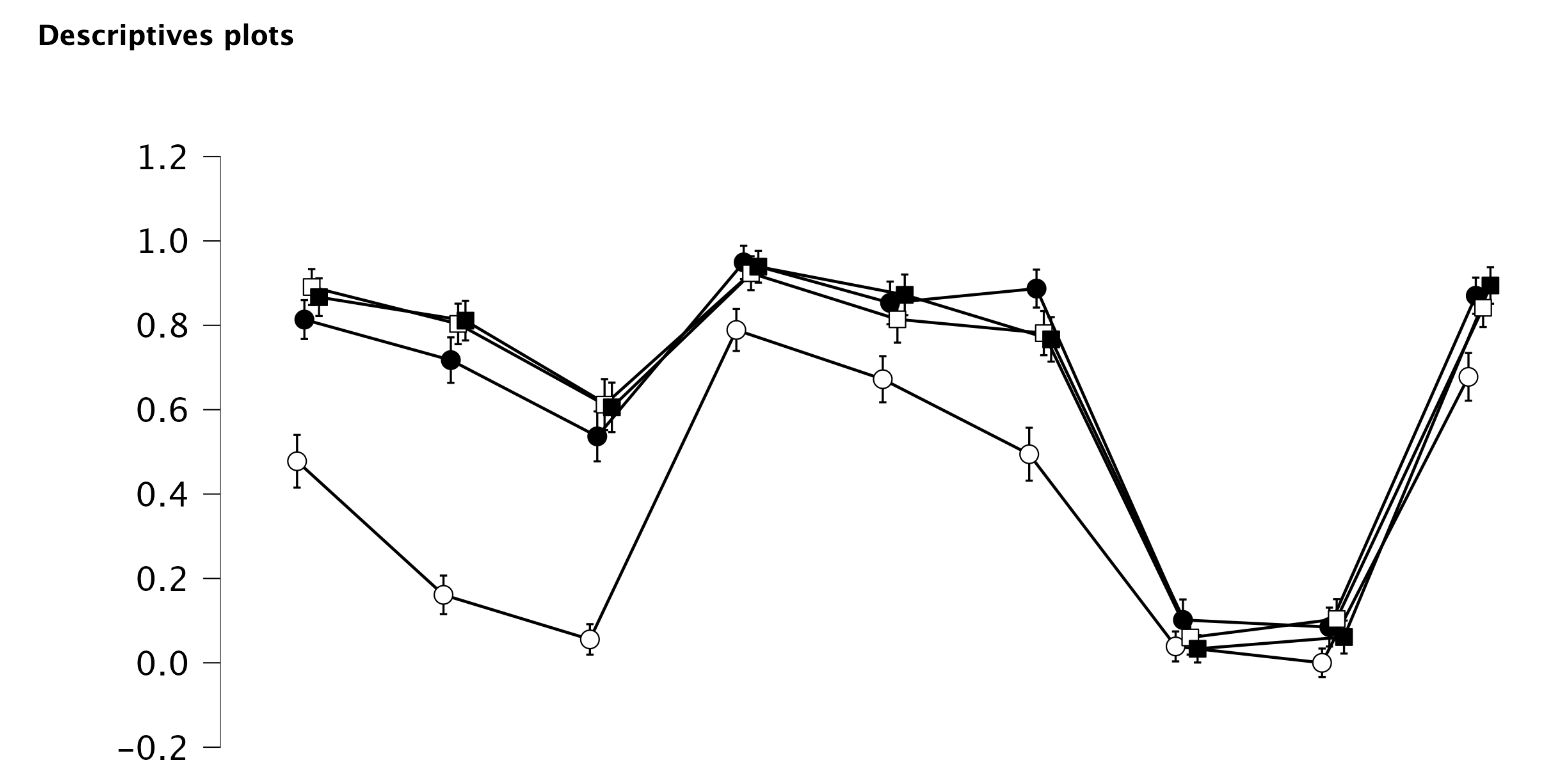} 
    \caption{\textbf{Descriptive Plots by Test Group.} Performance stratification across 9 items. The x-axis represents the items, and the y-axis represents the normalized score. One group (open circles) consistently underperforms compared to the others.}
    \label{fig:testgroup_plot}
\end{figure}

\noindent As illustrated in Figure \ref{fig:testgroup_plot}, while all groups struggled similarly on the most difficult items (e.g., Item 7), the lower-performing group (bottom line) showed significantly higher divergence on Items 1, 3, and 5 compared to the clustered higher-performing groups.

\section{Case Profiles}
\label{section: case-study}
\begin{figure}[H]
    \centering
    \includegraphics[width=0.95\textwidth]{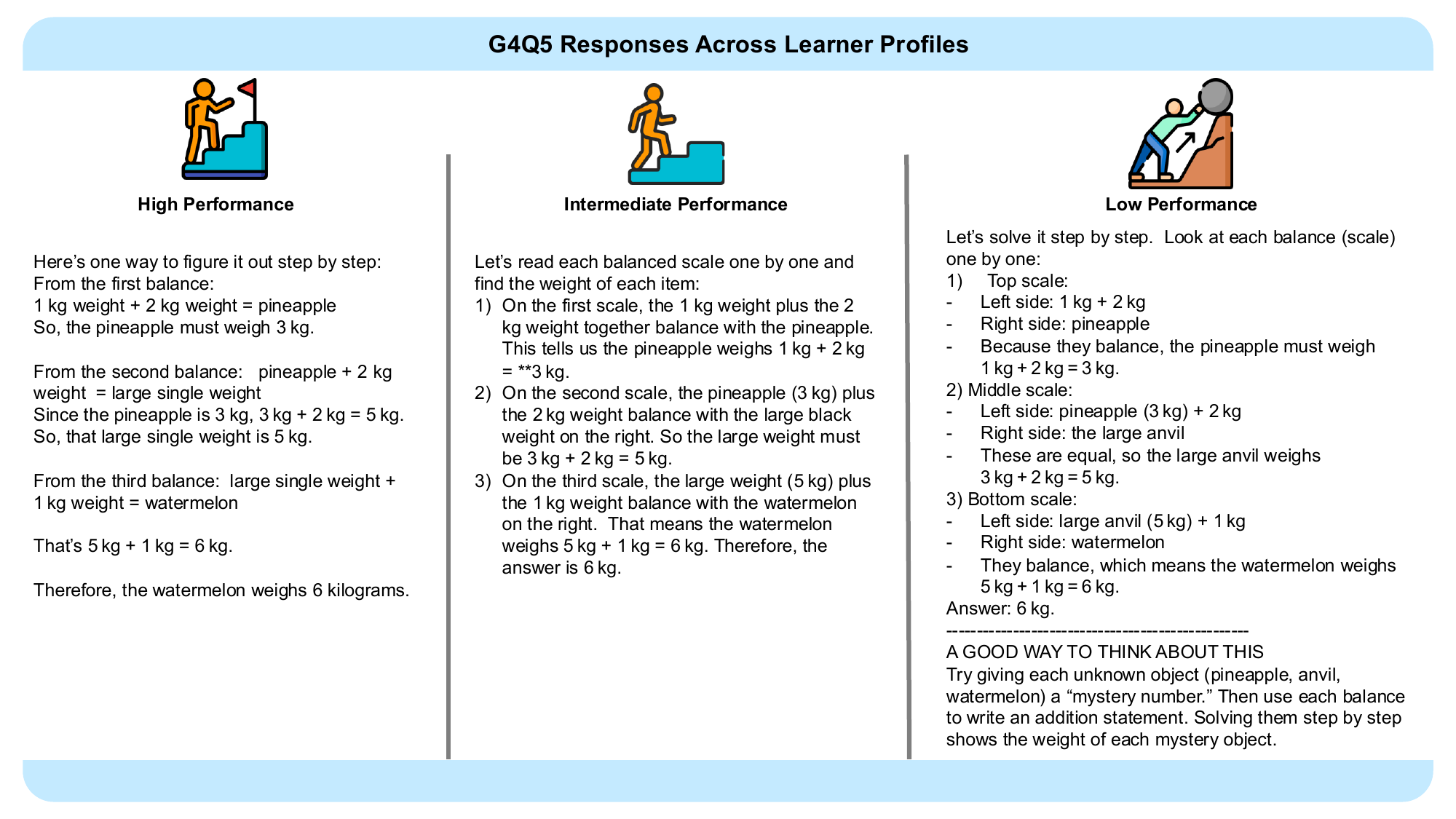}
    \caption{Case profiles for three learner profiles in G4Q5.}
    \label{figure: case-study}
\end{figure}

To qualitatively illustrate the adaptive capabilities of the evaluated models, we present a case study of a Grade 4 measurement problem. Figure \ref{figure: case-study} displays the responses generated by the model for the same math question across three distinct learner profiles: High-performance, Intermediate, and Low-performance. This visual comparison demonstrates how the model adjusts its pedagogical strategy, ranging from direct calculation to scaffolded instruction, based on the learner's stated confidence and prior knowledge.

\section{Example Responses Across Context Conditions}
\label{appendix:context_examples}

Table~\ref{tab:context_examples_o1} illustrates how an O1 model adapts its explanations across four context conditions (G1-G4) for a Grade 4 learner solving a balance scale problem. 

\begin{table}[h!]
\centering
\small
\begin{tabular}{p{0.05\linewidth} p{0.3\linewidth} p{0.55\linewidth}}
\toprule
\textbf{C} & \textbf{Prompt / Learner Profile} & \textbf{Model Response} \\
\midrule
G1 & Grade 4, low prior knowledge & A concise way to see this is: \\
    & & • From the first balance, the pineapple must weigh 1~kg + 2~kg = 3~kg. \\
    & & • The second balance shows that the 3~kg pineapple plus 2~kg equals the unlabeled weight, so that weight is 5~kg. \\
    & & • Finally, the third balance has that 5~kg weight plus 1~kg equals the watermelon’s mass. Hence, the watermelon weighs 6~kg. \\
\midrule
C2 & Grade 4, low prior knowledge + partial context & Here’s one way to think this through step-by-step: \\
    & & 1) Top scale: 1~kg + 2~kg = pineapple → 3~kg \\
    & & 2) Middle scale: pineapple (3~kg) + 2~kg = single heavy weight → 5~kg \\
    & & 3) Bottom scale: 5~kg + 1~kg = watermelon → 6~kg \\
    & & Tip: Writing equations for each scale helps solve the puzzle step by step. \\
\midrule
G3 & Grade 4, low prior knowledge + full context & Let’s solve it step by step: \\
    & & 1) Top scale: 1~kg + 2~kg = pineapple → 3~kg \\
    & & 2) Middle scale: pineapple (3~kg) + 2~kg = anvil → 5~kg \\
    & & 3) Bottom scale: anvil (5~kg) + 1~kg = watermelon → 6~kg \\
    & & GOOD WAY: Assign a “mystery number” to each object and solve each scale sequentially. \\
\midrule
G4 & Grade 4, low prior knowledge + full context + learner motivation & Let’s work through each scale, step by step: \\
    & & 1) Top: 1~kg + 2~kg = pineapple → 3~kg \\
    & & 2) Middle: pineapple (3~kg) + 2~kg = unlabeled weight → 5~kg \\
    & & 3) Bottom: 5~kg + 1~kg = watermelon → 6~kg \\
    & & Therefore, the watermelon weighs 6 kilograms. \\
\bottomrule
\end{tabular}
\caption{O1 model responses across context conditions (G1–G4) for a Grade 4 learner on a balance scale problem. Increasing context leads to more step-by-step guidance and scaffolding.}
\label{tab:context_examples_o1}
\end{table}

\section{Prompt Design Strategies}
\label{prompt design}

This section provides a comprehensive documentation of the evaluation rubric, presenting the specific criteria, scoring hierarchies, and illustrative examples utilized in the study. To ensure full transparency and reproducibility, the content below reproduces the definitions exactly as they appear in the evaluation framework. Each entry details the theoretical sub-dimension (e.g., Cognitive vs. Motivational aspects), the precise conditions required for each score (e.g., the distinction between a score of 0 and 1), and the reference examples that serve as ground truth for annotators.

\begin{table}[H]
\centering
\small
\renewcommand{\arraystretch}{1}
\begin{tabularx}{\textwidth}{L V C E}
\toprule
\textbf{Learner Profile Type} & \textbf{Grade} & \textbf{Prompt}  \\
\midrule


\textbf{}High-performance & 
\textbf{}4  & 
\textbf{}I am a student from Grade 4, I like learning mathematics very much and I am very confident in mathematics. Now I have mastered all mathematics topics in grade 4, including number, measurement and geometry, and data. I got 615 in the TIMSS 2019 Math Test. Can you teach me this math question? \\
\addlinespace

\textbf{}High-performance & 
\textbf{}8  & 
\textbf{}I am a student from Grade 8, I like learning mathematics very much and I am very confident in mathematics. Now I have mastered all mathematics topics in grade 8, including number, algebra, geometry and measurement, and data and probability. I got 625 in the TIMSS 2019 Math test. Can you teach me this math question? \\
\addlinespace

\textbf{}Intermediate-performance & 
\textbf{}4  & 
\textbf{}I am a student from Grade 4, I like learning mathematics very much and I am confident in mathematics. Now I have mastered all mathematics topics in grade 4, including number, measurement and geometry, and data. I got 550 in the TIMSS 2019 Math Test. Can you teach me this math question? \\
\addlinespace

\textbf{}Intermediate-performance & 
\textbf{}8  & 
\textbf{}I am a student from Grade 8, I like learning mathematics very much and I am very confident in mathematics. Now I have mastered all mathematics topics in grade 8, including number, algebra, geometry and measurement, and data and probability. I got 550 in the TIMSS 2019 Math test. Can you teach me this math question? \\
\addlinespace

\textbf{}Low-performance & 
\textbf{}4  & 
\textbf{}I am a student from Grade 4, I don’t like learning mathematics and I am not confident in mathematics. Now I have mastered mathematics topics such as number and data in grade 4. I got 390 in the TIMSS 2019 Math Test. Can you teach me this math question? \\
\addlinespace

\textbf{}Low-performance & 
\textbf{}8  & 
\textbf{}I am a student from Grade 8, I don’t like learning mathematics and I am not confident in mathematics. Now I have mastered mathematics topics such as number and geometry in grade 8. I got 390 in the TIMSS 2019 Math Test. Can you teach me this math question? \\

\bottomrule
\end{tabularx}
\caption{\textbf{User prompts.} User prompts across learner profiles (high-performance, intermediate-performance, and low-performance) and grade levels (grade 4 and 8).}
\label{tab:rubric_accuracy}
\end{table}

\clearpage

\section{Learning Goals \& Content Areas}
\label{learning goals}
\begin{table}[H]
\centering
\small
\renewcommand{\arraystretch}{1.4}

\begin{tabularx}{\textwidth}{L V C E}
\toprule
\textbf{Content Area}& \textbf{Topic} & \textbf{Learning Goal}  \\
\midrule

\multicolumn{3}{l}{\colorbox{gray!15}{\textbf{GRADE 4}}} \\
\addlinespace[0.5ex]

\textbf{}Number & 
\textbf{}Whole Numbers & 
\textbf{}1. Demonstrate knowledge of place value (2-digit to 6-digit numbers); represent whole numbers with words, diagrams, number lines, or symbols; order numbers.
\newline 
\textbf{}2. Add and subtract (up to 4-digit numbers), including computation in simple contextual problems.
\newline 
\textbf{}3. Multiply (up to 3-digit by 1-digit and 2-digit by 2-digit numbers) and divide (up to 3-digit by 1-digit numbers), including computation in simple contextual problems.
\newline 
\textbf{}4. Solve problems involving odd and even numbers, multiples and factors of numbers, rounding numbers (up to the nearest ten thousand), and making estimates.
\newline 
\textbf{}5. Combine two or more properties of numbers or operations to solve problems in context.
\\
\addlinespace

\textbf{} & 
\textbf{}Expressions, Simple Equations, and Relationships & 
\textbf{}1. Find the missing number or operation in a number sentence (e.g., 17 + w = 29).
\newline 
\textbf{}2. Identify or write expressions or number sentences to represent problem situations that may involve unknowns.
\newline 
\textbf{}3. Identify and use relationships in a well-defined pattern (e.g., describe the relationship between adjacent terms and generate pairs of whole numbers given a rule).
\\
\addlinespace

\textbf{} & 
\textbf{}Fractions and Decimals & 
\textbf{}1. Recognize fractions as parts of wholes or collections; represent fractions using words, numbers, or models; compare and order simple fractions; add and subtract simple fractions, including those set in problem situations. (Fractions may have denominators of 2, 3, 4, 5, 6, 8, 10, 12, or 100.)
\newline 
\textbf{}2. Demonstrate knowledge of decimal place value including representing decimals using words, numbers, or models; compare, order, and round decimals; add and subtract decimals, including those set in problem situations. (Decimals may have one or two decimal places, allowing for computations with money.)
\\
\addlinespace

\textbf{}Measurement and Geometry & 
\textbf{}Measurement & 
\textbf{}1. Measure and estimate lengths (millimeters, centimeters, meters, kilometers); solve problems involving lengths.
\newline 
\textbf{}2. Solve problems involving mass (gram and kilogram), volume (milliliter and liter), and time (minutes and hours); identify appropriate types and sizes of units and read scales.
\newline 
\textbf{}3. Solve problems involving perimeters of polygons, areas of rectangles, areas of shapes covered with squares or partial squares, and volumes filled with cubes.
\\
\addlinespace

\textbf{} & 
\textbf{}Geometry & 
\textbf{}1. Identify and draw parallel and perpendicular lines; identify and draw right angles and angles smaller or larger than a right angle; compare angles by size.
\newline 
\textbf{}2. Use elementary properties, including line and rotational symmetry, to describe, compare, and create common two-dimensional shapes (circles, triangles, quadrilaterals, and other polygons).
\newline 
\textbf{}3. Use elementary properties to describe and compare three-dimensional shapes (cubes, rectangular solids, cones, cylinders, and spheres) and relate these with their two-dimensional representations.
\\
\addlinespace

\textbf{}Data & 
\textbf{}Reading, Interpreting, and Representing Data & 
\textbf{}1. Read and interpret data from tables, pictographs, bar graphs, line graphs, and pie charts.
\newline 
\textbf{}2. Organize and represent data to help answer questions.
\\
\addlinespace

\textbf{}Data & 
\textbf{}Using Data to Solve Problems & 
\textbf{}Use data to answer questions that go beyond directly reading data displays (e.g., solve problems and perform computations using data, combine data from two or more sources, draw conclusions based on data).
\\
\bottomrule
\end{tabularx}
\caption{\textbf{Content Areas \& Learning Goals.} TIMSS mathematical content areas and learning goals in Grade 4.}
\label{tab:rubric_accuracy}

\end{table}

\clearpage

\begin{table}[H]
\centering
\small
\renewcommand{\arraystretch}{1.4}

\begin{tabularx}{\textwidth}{L V C E}
\toprule
\textbf{Content Area}& \textbf{Topic} & \textbf{Learning Goal}  \\
\midrule

\multicolumn{3}{l}{\colorbox{gray!15}{\textbf{GRADE 8}}} \\
\addlinespace[0.5ex]

\textbf{}Number & 
\textbf{}Integers & 
\textbf{}1. Demonstrate understanding of properties of numbers and operations; find and use multiples and factors, identify prime numbers, evaluate positive integer powers of numbers, evaluate square roots of perfect squares up to 144, and solve problems involving square roots of whole numbers.
\newline 
\textbf{}2. Compute and solve problems with positive and negative numbers, including through movement on the number line or various models (e.g., losses and gains, thermometers).
\\
\addlinespace

\textbf{} & 
\textbf{}Fractions and Decimals & 
\textbf{}1. Using various models and representations, compare and order fractions and decimals, and identify equivalent fractions and decimals.
\newline 
\textbf{}2. Compute with fractions and decimals, including those set in problem situations.
\\
\addlinespace

\textbf{} & 
\textbf{}Ratio, Proportion, and Percent & 
\textbf{}1. Identify and find equivalent ratios; model a given situation by using a ratio; divide a quantity according to a given ratio.
\newline 
\textbf{}2. Solve problems involving proportions or percents, including converting between percents and fractions or decimals.
\\
\addlinespace

\textbf{}Algebra & 
\textbf{}Expressions, Operations, and Equations & 
\textbf{}1. Find the value of an expression or a formula given values of the variables.
\newline 
\textbf{}2. Simplify algebraic expressions involving sums, products, and powers; compare expressions to determine if they are equivalent.
\newline 
\textbf{}3. Write expressions, equations, or inequalities to represent problem situations.
\newline 
\textbf{}4. Solve linear equations, linear inequalities, and simultaneous linear equations in two variables, including those that model real life situations.
\\
\addlinespace

\textbf{} & 
\textbf{}Relationships and Functions & 
\textbf{}1. Interpret, relate and generate representations of linear functions in tables, graphs, or words; identify properties of linear functions including slope and intercepts.
\newline 
\textbf{}2. Interpret, relate and generate representations of simple non-linear functions (e.g., quadratic) in tables, graphs, or words; generalize pattern relationships in a sequence using numbers, words, or algebraic expressions.
\\
\addlinespace

\textbf{}Geometry & 
\textbf{}Geometric Shapes and Measurements & 
\textbf{}1. Identify and draw types of angles and pairs of lines and use the relationships between angles on lines and in geometric figures to solve problems, including those involving the measures of angles and line segments; solve problems involving points in the Cartesian plane.
\newline 
\textbf{}2. Identify two-dimensional shapes and use their geometric properties to solve problems, including those involving perimeter, circumference, area, and the Pythagorean Theorem.
\newline 
\textbf{}3. Recognize and draw images of geometric transformations (translations, reflections, and rotations) in the plane; identify congruent and similar triangles and rectangles and solve related problems.
\newline 
\textbf{}4. Identify three-dimensional shapes and use their geometric properties to solve problems, including those involving surface area and volume; relate three-dimensional shapes with their two-dimensional representations.
\\
\addlinespace

\textbf{}Data and Probability & 
\textbf{}Data & 
\textbf{}1. Read and interpret data from one or more sources to solve problems (e.g., interpolate and extrapolate, make comparisons, draw conclusions).
\newline 
\textbf{}2. Identify appropriate procedures for collecting data; organize and represent data to help answer questions.
\newline 
\textbf{}3. Calculate, use, or interpret statistics (i.e., mean, median, mode, range) summarizing data distributions; recognize the effect of spread and outliers.
\\
\addlinespace

\textbf{} & 
\textbf{}Probability & 
\textbf{}For simple and compound events: a) determine theoretical probability (based on equally likely outcomes, e.g., rolling a fair die) or b) estimate the empirical probability (based on experimental outcomes).
\\

\bottomrule
\end{tabularx}
\caption{\textbf{Content Areas \& Learning Goals.} TIMSS mathematical content areas and learning goals in Grade 8.}
\label{tab:rubric_accuracy}
\end{table}

\end{document}